\documentclass[10pt,twocolumn,letterpaper]{article}

\usepackage[pagenumbers]{cvpr} 
\usepackage{tikz}
\usepackage{lipsum}
\usepackage{pgfplots}
\usepackage{pgfplotstable}
\usepackage{multirow}
 
\usepackage{algorithm}
\usepackage{listings}
\lstdefinestyle{mocov3}{
  backgroundcolor=\color{white},
  basicstyle=\fontsize{7.5pt}{7.5pt}\ttfamily\selectfont,
  columns=fullflexible,
  breaklines=true,
  captionpos=b,
  commentstyle=\fontsize{7.5pt}{7.5pt}\color[rgb]{0.25,0.5,0.5},
  keywordstyle=\fontsize{7.5pt}{7.5pt}\color[rgb]{0.85,0.18,0.50},
}
\usepackage{listings}

\usepackage[hang]{footmisc}  
\newcommand\blfootnote[1]{%
    \begingroup
    \renewcommand\thefootnote{}\footnote{#1}%
    \addtocounter{footnote}{-1}%
    \endgroup
}

\newcommand{\rev}{}
\newcommand{\cam}{}

\definecolor{cvprblue}{rgb}{0.21,0.49,0.74}
\usepackage[pagebackref,breaklinks,colorlinks,allcolors=cvprblue]{hyperref}

\def\paperID{14625} 
\def\confName{CVPR}
\def\confYear{2025}

\usepackage{CJKutf8}

\title{StarGen: A Spatiotemporal Autoregression Framework with Video Diffusion Model for Scalable and Controllable Scene Generation}

\author{%
Shangjin Zhai$^{1}$$^{\ast}$,
Zhichao Ye$^{1}$$^{\ast}$,
Jialin Liu$^{1}$,
Weijian Xie$^{1}$,
Jiaqi Hu$^{1}$,
Zhen Peng$^{1}$,
Hua Xue$^{1}$,\\
Danpeng Chen$^{2,3}$,
Xiaomeng Wang$^{1}$,
Lei Yang$^{1}$,
Nan Wang$^{1}$,
Haomin Liu$^{1}$$^{\dagger}$,
Guofeng Zhang$^{2}$$^{\dagger}$
\\\\
$^{1}$SenseTime Research~~~
$^{2}$State Key Lab of CAD\&CG, Zhejiang University~~~
$^{3}$Tetras.AI
}

\begin{document}

\twocolumn[{%
\renewcommand\twocolumn[1][]{#1}%
\maketitle

\vspace{-2em}
\centering
\includegraphics[width=0.9\linewidth]{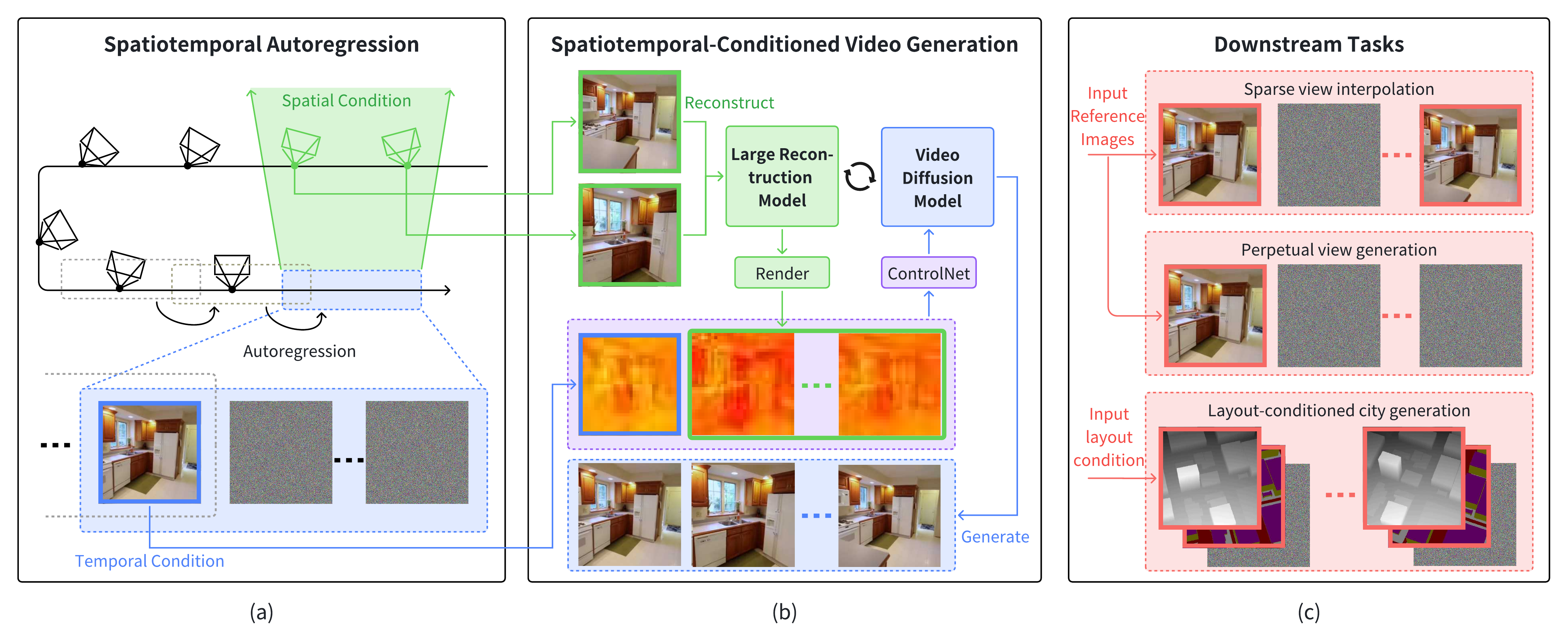}
\captionof{figure}{
Overview of the proposed method:
(a) We introduce a spatiotemporal autoregression framework for long-range scene generation.
The generated scene is represented as a set of sparsely sampled posed images.
The generation of the current sliding window of images (blue dotted box) is conditioned on spatially adjacent images (green frustums) and temporally overlapping image (blue solid box).
(b) Spatial conditioning images are processed by a large reconstruction model, which extracts the 3D information and renders the reconstructed latent features to each novel view.
These spatial features, together with the temporal conditioning image, are used to condition the generation of the current window through a video diffusion model and a ControlNet.
(c) The framework is used to implement three downstream tasks, including sparse view interpolation, perpetual view generation, and layout-conditioned city generation.
} 
\vspace{1em}
\label{fig:teaser}
}]


\blfootnote{$^{\ast}$ indicates equal contribution.}
\blfootnote{$^{\dagger}$indicates corresponding author.}
\blfootnote{ This work was partially supported by the National Key Research and Development Program of China (No. 2023YFF0905104) and NSF of China (No. 62425209).}

\begin{abstract}
Recent advances in large reconstruction and generative models have significantly improved scene reconstruction and novel view generation. However, due to compute limitations, each inference with these large models is confined to a small area, making long-range consistent scene generation challenging. To address this, we propose StarGen, a novel framework that employs a pre-trained video diffusion model in an autoregressive manner for long-range scene generation. \rev{The generation of each video clip} is conditioned on the 3D warping of spatially adjacent images and the temporally overlapping image from \rev{previously generated clips}, improving spatiotemporal consistency in long-range scene generation with precise pose control. The spatiotemporal condition is compatible with various input conditions, facilitating diverse tasks, including sparse view interpolation, perpetual view generation, and layout-conditioned city generation. Quantitative and qualitative evaluations demonstrate StarGen's superior scalability, fidelity, and pose accuracy compared to state-of-the-art methods. Project page: \url{https://zju3dv.github.io/StarGen}.

\end{abstract}    

\vspace{-2em}
\section{Introduction}
\label{sec:intro}

\rev{
In recent years, the rapid development of large models has driven significant progress in 3D reconstruction and generation, with the two fields increasingly intertwining and complementing each other.
On the reconstruction side, the emergence of large reconstruction models~\cite{DBLP:conf/cvpr/YuYTK21,DBLP:conf/iclr/Hong0GBZLLSB024,DBLP:conf/eccv/ZhangBTXZSX24} has successfully reduced the reliance on dense multi-view captures.
Meanwhile, generation models have been leveraged to address the challenge of filling invisible regions in sparsely captured views~\cite{DBLP:conf/cvpr/WuMHPGWSVBPH24,DBLP:journals/corr/abs-2408-16767,DBLP:journals/corr/abs-2409-02048}.
On the generation side, 3D reconstruction techniques have facilitated the lifting of 2D generation models to 3D generation tasks, either by distilling 2D distributions into a 3D representation~\cite{DBLP:conf/iclr/PooleJBM23,DBLP:conf/nips/Wang00BL0023,DBLP:conf/cvpr/LiangYLLXC24}, or by reconstructing the 3D representation from 2D generated images~\cite{DBLP:conf/iccv/HolleinCO0N23,DBLP:journals/tvcg/ZhangLWWL24,DBLP:journals/corr/abs-2311-13384}.
In addition, large reconstruction models have also been utilized to enhance consistency among generated 2D multi-view images~\cite{DBLP:journals/corr/abs-2403-12010,DBLP:journals/corr/abs-2409-02048,DBLP:journals/corr/abs-2405-10314}.
}

One major limitation of these large reconstruction and generation models is that, under limited compute resources, only a restricted number of tokens can be processed in a single inference. 
\rev{Consequently, current methods are typically confined to generating single objects or short-range scenes, making it challenging to support long-range scene generation.}
While there are temporal autoregression methods that condition the first frames of a current video clip on the last frames of the previously generated clip to enable long-range video generation, they only maintain temporal consistency over short periods~\cite{DBLP:conf/siggraph/Deng0LGSW24,DBLP:journals/corr/abs-2405-17398}. As time progresses, errors accumulate, and spatial consistency becomes difficult to preserve. For example, when moving back and forth within the same area, each pass may generate inconsistent content at the same location.

In this work, we present \textbf{StarGen}, a \textbf{S}patio\textbf{T}emporal \textbf{A}uto\textbf{R}egression framework for long-range scene \textbf{Gen}eration. As illustrated in~\cref{fig:teaser}, the key idea is to condition the generation of each video clip not only on temporally adjacent images but also on spatially adjacent ones that share common content with the current window. We introduce a large reconstruction model that extracts 3D information from the spatial conditioning images and renders the reconstructed latent features to each novel view, guiding the generation with precise pose control. Our contributions are summarized as follows:
\begin{itemize}
    \item We propose StarGen, a novel autoregression framework that combines both spatial and temporal conditions to support long-range scene generation with precise pose control.
    \item We present a novel architecture that complements a carefully designed large reconstruction model with a pre-trained video diffusion model for spatiotemporal conditioned video generation.
    \item We demonstrate the versatility of StarGen by implementing three tasks within the framework, including sparse view interpolation, perpetual view generation, and layout-conditioned city generation.
    \item We conduct quantitative and qualitative evaluations demonstrating that StarGen achieves superior scalability, fidelity, and pose accuracy compared to state-of-the-art methods.
\end{itemize}
\section{Related Work}
\label{sec:relatedwork}





\noindent\textbf{The Reconstruction Models.}
The traditional 3D reconstruction pipeline includes Structure from Motion for camera pose estimation~\cite{DBLP:journals/tip/ZhangLDJWB16,DBLP:journals/tcsv/YeBZLBZ24}, Multi-View Stereo for point cloud reconstruction~\cite{DBLP:conf/iccv/GoeseleSCHS07,DBLP:conf/iccv/ChenHXS19}, and mesh extraction with texturing for novel view synthesis~\cite{DBLP:conf/ismar/NewcombeIHMKDKSHF11,DBLP:conf/cvpr/ChoiSL0S00MA024}. 
Recent advances, like neural radiance fields (NeRF)~\cite{DBLP:journals/cacm/MildenhallSTBRN22,DBLP:conf/cvpr/BarronMVSH22,DBLP:conf/iccv/BarronMVSH23,DBLP:conf/eccv/ChenXGYS22}, use MLPs to represent geometry and appearance implicitly, 
\rev{significantly improving novel view synthesis quality but at the cost of longer rendering times.}
The 3D-GS~\cite{DBLP:journals/tog/KerblKLD23} method, which uses 3D Gaussian point clouds and efficient Gaussian splatting, significantly improves the rendering efficiency.
\rev{Further research~\cite{DBLP:conf/cvpr/YuCHS024,DBLP:conf/cvpr/0005YXX0L024,DBLP:conf/siggraph/HuangYC0G24,DBLP:journals/corr/abs-2406-06521} has enhanced its rendering quality and geometric accuracy, while also extending its application to dynamic scenes~\cite{DBLP:conf/cvpr/ZhouSXBQL00L24,DBLP:conf/cvpr/LiCLX24}. However, dense multi-view captures are still required, which limits broader applicability.} To address this, recent work has focused on feed-forward regression models for sparse view reconstruction. PixelNeRF~\cite{DBLP:conf/cvpr/YuYTK21} pioneered this approach by regressing pixel-aligned neural radiance fields, and later NeRF-based methods have enhanced feature matching~\cite{DBLP:conf/iccv/ChenXZZXY021}, 3D representation~\cite{DBLP:conf/cvpr/XuCCSZP0024, DBLP:conf/iccv/ChenXZZXY021}, and \rev{model architecture design}~\cite{DBLP:conf/cvpr/DuSTS23, DBLP:conf/cvpr/SajjadiMPBGRVLD22}. \rev{Concurrently, methods such as SplatterImage~\cite{DBLP:conf/cvpr/Szymanowicz0V24}, PixelSplat~\cite{DBLP:conf/cvpr/CharatanLTS24}, and similar works~\cite{DBLP:journals/corr/abs-2403-14627, xu2024depthsplat, DBLP:journals/corr/abs-2405-17958} adopt 3D-GS as a lightweight alternative to NeRF. More recently, transformer-based architectures~\cite{DBLP:conf/iclr/Hong0GBZLLSB024, DBLP:conf/eccv/ZhangBTXZSX24, DBLP:journals/corr/abs-2403-14621, DBLP:conf/eccv/TangCCWZL24} have been introduced, leveraging their scalability. Despite these advancements, the reconstruction of occluded or invisible regions from sparse views remains a significant challenge.}

\noindent\textbf{The Generation Models.}
\rev{
The early era of generation models was dominated by Generative Adversarial Networks (GANs), which learn to generate data through an adversarial process between a generator and a discriminator~\cite{DBLP:conf/nips/GoodfellowPMXWOCB14,DBLP:conf/iccv/ZhuPIE17,DBLP:conf/cvpr/KarrasLA19}.
Later, diffusion models emerged and demonstrated superior performance over GANs in a wide range of tasks, leveraging a gradual denoising process that transforms random noise into high-quality samples~\cite{DBLP:conf/nips/SongE19,DBLP:conf/nips/HoJA20,DBLP:conf/cvpr/RombachBLEO22}.
Recently, Diffusion Transformers (DiT)~\cite{DBLP:conf/iccv/PeeblesX23,DBLP:journals/corr/abs-2401-08740,DBLP:conf/iclr/ChenYGYXWK0LL24} have established themselves as a more powerful alternative to traditional UNet-based architectures, becoming the mainstream backbone for diffusion models due to their scalability and ability to model long-range dependencies.
ControlNet-like mechanisms~\cite{DBLP:conf/iccv/ZhangRA23,DBLP:journals/corr/abs-2401-05252} further improve the controllability of diffusion models through conditional inputs, allowing for fine-grained control over generated content.
These advances have been expanded into video generation by training on video data, enabling the creation of coherent image sequences~\cite{DBLP:conf/iclr/0002YRL00AL024,DBLP:journals/corr/abs-2311-15127,DBLP:journals/corr/abs-2408-06072}.
Recent works~\cite{DBLP:conf/siggraph/WangYWLCXLS24,DBLP:journals/corr/abs-2404-02101} have also introduced motion control mechanisms to guide the temporal dynamics of generated videos, significantly enhancing their practicality for real-world applications.
The 2D diffusion models have also be lifted to 3D generation tasks.
Early methods achieve this by distilling 2D distributions into 3D representations~\cite{DBLP:conf/iclr/PooleJBM23,DBLP:conf/cvpr/Lin0TTZHKF0L23,DBLP:conf/iccv/LiuWHTZV23}, but they are typically limited to generating single objects.
Other approaches attempt to generate larger scenes through incremental inpainting~\cite{DBLP:conf/iccv/HolleinCO0N23,DBLP:conf/nips/FridmanAKD23,DBLP:conf/cvpr/LiangYLLXC24}.
However, due to the lack of spatial constraints between different generation steps, they often suffer from poor long-term consistency.
}

\noindent\textbf{Combining Reconstruction and Generation.}
Nowadays, the task of reconstruction and generation are beginning to merge, gradually alleviating the limitations of each individual task~\cite{DBLP:conf/cvpr/WuMHPGWSVBPH24,DBLP:journals/corr/abs-2408-16767,DBLP:journals/corr/abs-2409-02048}. These methods use large reconstruction models~\cite{DBLP:conf/cvpr/YuYTK21,DBLP:conf/cvpr/Wang0CCR24} to reconstruct visible regions while employing diffusion models~\cite{DBLP:conf/cvpr/RombachBLEO22,DBLP:conf/eccv/XingXZCYLLWSW24} to fill in the invisible areas, enabling sparse view reconstruction and even perpetual view generation. Specifically, ReconFusion~\cite{DBLP:conf/cvpr/WuMHPGWSVBPH24} uses PixelNeRF~\cite{DBLP:conf/cvpr/YuYTK21} for reconstruction of visible regions, and image diffusion~\cite{DBLP:conf/cvpr/RombachBLEO22} for generation of invisible regions. However, since the images are generated independently, inconsistencies can arise between consecutive frames. NeRF reconstruction is used to mitigate these inconsistencies by averaging them out, though it can introduce some blurriness. Concurrent works ReconX~\cite{DBLP:journals/corr/abs-2408-16767} and ViewCrafter~\cite{DBLP:journals/corr/abs-2409-02048} improve temporal consistency through video diffusion models. However, under limited compute resources, only a short clip of images can be processed by a video diffusion model in a single inference. As a result, only intra-clip consistency can be guaranteed. To improve inter-clip consistency, ReconX treats adjacent sparse input images as the first and last frames of a generated clip. It employs DUSt3R~\cite{DBLP:conf/cvpr/Wang0CCR24} to reconstruct the point cloud from the two images, which is encoded as a condition for the next clip generation. Similarly, ViewCrafter uses point cloud as a global representation, projecting the previously reconstructed point cloud onto the current clip as the condition. After generating current clip, the generated images are used to reconstruct and expand the global point cloud, enabling perpetual view generation in the autoregressive manner. However, due to inherent errors in point cloud reconstruction, which accumulate from one clip to the next, scalability is limited. In contrast, the proposed StarGen samples generated images as the global representation
to mitigate error accumulation and improve scalability.

\vspace{-1em}
\section{Method}
\label{sec:method}

\begin{figure*}[htbp]
  \centering
  \includegraphics[width=0.9\textwidth]{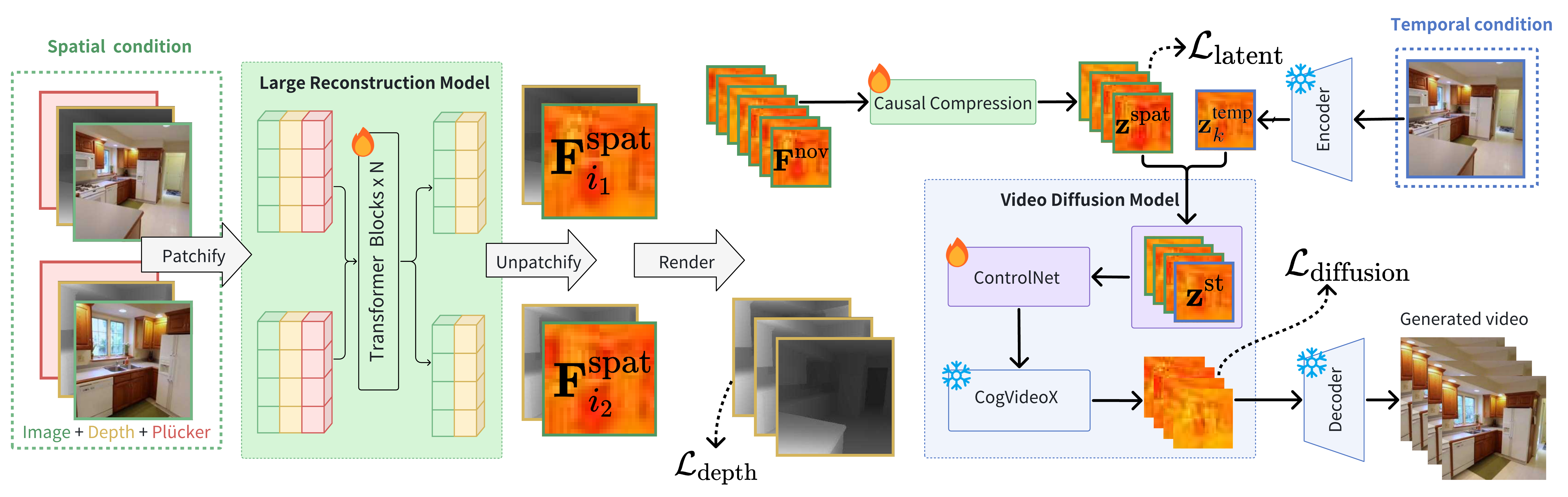}
  \caption{
  Spatiotemporal-Conditioned Video Generation. Given two posed images as spatial conditions \rev{(green dotted box on the left)}, the large reconstruction model regresses their depth maps and feature maps. \rev{The two feature maps $\mathbf{F}_{i_1}^\text{spat}$ and $\mathbf{F}_{i_2}^\text{spat}$ are rendered into novel views $\mathbf{F}^\text{nov}$ and temporally compressed to the latent space of CogVideoX, resulting in $\mathbf{z}^\text{spat}$. Simultaneously, the temporal conditioning image (blue dotted box on the right) is encoded to $\mathbf{z}^\text{temp}_k$ to replace the corresponding latent in $\mathbf{z}^\text{spat}$, resulting in the spatiotemporal condition $\mathbf{z}^\text{st}$, which conditions the generation of CogVideoX through a ControlNet.}
  } 
  \label{fig:pipeline} 
  \vspace{-1em}
\end{figure*}

An overview of the proposed method is presented in~\cref{fig:teaser}. The approach is an autoregression framework for long-range scene generation, where each step generates a sliding window of images conditioned on the previously generated content, as illustrated in~\cref{fig:teaser}(a) and explained in~\cref{sec:method-ar}. Within each step, the method for generating the current sliding window is shown in~\cref{fig:teaser}(b) and detailed in~\cref{sec:method-gen}. The framework is used to implement three downstream tasks as depicted in~\cref{fig:teaser}(c) and described in~\cref{sec:method-task}.

\subsection{Spatiotemporal Autoregression}
\label{sec:method-ar}

Given a long-range pose trajectory, StarGen generates an image for each pose. 
Similar to previous temporal autoregression methods based on video diffusion models~\cite{DBLP:conf/siggraph/Deng0LGSW24,DBLP:journals/corr/abs-2405-17398}, the long-range generation progresses through overlapping sliding windows $\{{\mathbf{W}_k}\}^K_{k=1}$ of short video clips, indicated by the dotted boxes in~\cref{fig:teaser}(a). 
The generation of current window ${\mathbf{W}_k}$ is conditioned on the temporally overlapping image ${\mathbf{I}^\text{temp}_k}$ from the previous window ${\mathbf{W}_{k-1}}$, as indicated by the blue \rev{solid} box. 
To handle cases where temporally non-adjacent images may share \rev{common} content \rev{with current window}, we introduce spatially adjacent images to enhance spatial consistency.
Specifically, we represent the generated scene as a set of sparsely sampled images \rev{$\{{\mathbf{I}_i^\text{spat}}\}_{i=1}^M$}, each paired with an input pose ${\mathbf{P}_i}$ and a generated depth map ${\mathbf{D}_i}$. We identify the two \rev{spatial conditioning} images $({\mathbf{I}_{i_1}^\text{spat}},{\mathbf{I}_{i_2}^\text{spat}})$ with the largest common area with the current window, illustrated by the green frustums. These spatial and temporal conditioning images $({\mathbf{I}^\text{spat}_{i_1}},{\mathbf{I}^\text{spat}_{i_2}},{\mathbf{I}^\text{temp}_k})$ are fed into a spatiotemproal-conditioned video generation model to generate the current clip, which is illustrated in~\cref{fig:teaser}(b) and will be
detailed in~\cref{sec:method-gen}. Finally, two images are evenly sampled from the current clip and added to the set of sampled images, and the window slides forward.

\subsection{Spatiotemporal-Conditioned Video Generation}
\label{sec:method-gen}


\rev{
We carefully design a large reconstruction model that is combined with a pretrained video diffusion model to generate a video clip conditioned on the previously generated content, as illustrated in \cref{fig:pipeline}.
Formally, given the spatial condition $\mathbf{C}^\text{spat} = \{\mathbf{I}^\text{spat}_i, \mathbf{P}_i\}_{i=i_1,i_2}$, the temporal condition $\mathbf{C}^\text{temp} = \{\mathbf{I}^\text{temp}_k\}$, a text prompt $\mathbf{T}$, and the novel view poses $\mathbf{P}^\text{nov} = \{\mathbf{P}_j\}_{j=1}^N$, the goal is to model the conditional distribution of generated images for novel views $\mathbf{x} = \{\mathbf{x}_j\}_{j=1}^N$:
\begin{equation}
\mathbf{x} \sim p_\theta(\mathbf{x} | \mathbf{C}^\text{spat}, \mathbf{C}^\text{temp}, \mathbf{T}, \mathbf{P}^\text{nov}),
\end{equation}
where \( \theta \) represents the model parameters, and $N$ is the number of images in each video clip. This formulation enables the model to leverage both spatial and temporal conditions to guide the generation of novel view images.
}

\rev{
\noindent\textbf{Spatial Condition.}
Inspired by latenSplat~\cite{DBLP:conf/eccv/WewerRISL24}, we predict the scene structure from the two spatial conditioning images, and render the reconstructed latent features for each novel view to guide the generation.
Specifically, we utilize a Large Reconstruction Model (LRM) to
predict scene structure from two spatial conditioning images $({\mathbf{I}_{i_1}^\text{spat}},{\mathbf{I}_{i_2}^\text{spat}})$. Unlike existing methods~\cite{DBLP:journals/corr/abs-2403-14627,DBLP:journals/corr/abs-2405-17958,DBLP:conf/cvpr/CharatanLTS24}, which rely solely on color information, our approach incorporates a prior depth map and Plücker coordinates~\cite{plucker1865xvii} as additional inputs.
The complete LRM input is $\{\mathbf{I}^\text{spat}_i,\hat{\mathbf{D}}^\text{spat}_i,\hat{\textbf{P}}^\text{spat}_i\}_{i=i_1,i_2}$, including the RGB image $\mathbf{I}^\text{spat}_i\in \mathbb{R}^{H \times W \times 3}$, the depth map $\hat{\mathbf{D}}^\text{spat}_i \in \mathbb{R}^{H\times W\times 1}$ predicted by Depth Anything V2~\cite{DBLP:journals/corr/abs-2406-09414}, and the Plücker coordinates $ \hat{\mathbf{P}}^\text{spat}_i \in \mathbb{R}^{H\times W \times 6}$ derived from the input pose ${\mathbf{P}_i}$, where $H$ and $W$ are image height and width respectively. Following GS-LRM~\cite{DBLP:conf/eccv/ZhangBTXZSX24}, we stack, patchify, and concatenate the LRM inputs into a sequence of tokens.
These tokens are fed into a transformer network to regress the depth maps $\{\mathbf{D}_i^\text{spat}\}_{i=i_1,i_2}$ and feature maps $\{\mathbf{F}_i^\text{spat}\}_{i=i_1,i_2}$ corresponding to the two conditioning views.
Note that, unlike the scale-free $\hat{\mathbf{D}}_i^\text{spat}$ obtained from monocular depth prediction, the regressed $\mathbf{D}_i^\text{spat}$ is expected to be scale-aligned with the input poses.
So we can render the features into novel views given the novel view poses $\mathbf{P}^\text{nov} = \{\mathbf{P}_j\}_{j=1}^N$, obtaining the novel view features $\mathbf{F}^\text{nov} = \{\mathbf{F}_j^\text{nov}\}_{j=1}^N$ and depth maps $\mathbf{D}^\text{nov} = \{\mathbf{D}_j^\text{nov}\}_{j=1}^N$:
\begin{equation}
\mathbf{F}^\text{nov}, \mathbf{D}^\text{nov}=\mathcal{R}(\{\mathbf{F}^\text{spat}_i,\mathbf{D}^\text{spat}_i,\mathbf{P}_i\}_{i=i_1,i_2},\mathbf{P}^\text{nov}).
\end{equation}
The invisible regions in $\mathbf{F}^\text{nov}$ are filled with zeros. In this way, the structure, appearance, camera motion, and visibility information are encoded in $\mathbf{F}^\text{nov}$, which will condition the generation process to ensure that the visible regions are consistent with the spatial conditioning images, the invisible regions are identified and inpainted, and the input pose trajectory is followed. To align the condition to the latent space of video diffusion model, we set the dimension of each feature map to match the latent dimension, i.e. $\mathbf{F}_j^\text{nov}\in\mathbb{R}^{h\times w\times c}$, and train a small causal compression network to compress the temporal dimension of $\mathbf{F}^\text{nov}\in\mathbb{R}^{N\times h\times w\times c}$ from $N$ to $n$, resulting in $\mathbf{z}^\text{spat}\in\mathbb{R}^{n\times h\times w\times c}$. Following CogVideoX~\cite{DBLP:journals/corr/abs-2408-06072}, every four frames except the first one are compressed into one frame, resulting in $n=1+(N-1)/4$.
}

\rev{
\noindent\textbf{Spatiotemporal Condition.} Previous methods typically support temporal conditioning by tuning the backbone T2V model into an I2V model~\cite{DBLP:journals/corr/abs-2408-06072}, resulting in the need to train different backbones for conditioning at different positions~\cite{DBLP:conf/eccv/XingXZCYLLWSW24}. To enhance the versatility without backbone tuning, we directly replace the corresponding latent frame in $\mathbf{z}^\text{spat}$. Specifically, the temporal conditioning image $\mathbf{I}_k^\text{temp}$ is fed into the pretrained VAE encoder to obtain a latent $\mathbf{z}_k^\text{temp}$. For conditioning on the first frame, the first latent frame in $\mathbf{z}^\text{spat}$ is directly replaced with $\mathbf{z}_k^\text{temp}$. For conditioning on other frame, we modify the novel view pose trajectory by inserting three duplicated poses at frame $k$. These four static frames will be compressed into one in $\mathbf{z}^\text{spat}$, which is replaced with $\mathbf{z}_k^\text{temp}$. The replaced latent is denoted as $\mathbf{z}^\text{st}$, which serves as the condition for CogVideoX through a ControlNet~\cite{DBLP:journals/corr/abs-2401-05252} to generate a video consistent with the spatial and temporal conditions.
}

\rev{
\noindent\textbf{Training Loss.} The training loss combines three terms:
\begin{equation}
  \mathcal{L} = \lambda_{\text{depth}}\mathcal{L}_{\text{depth}} + \lambda_{\text{latent}}\mathcal{L}_{\text{latent}} + \lambda_{\text{diffusion}}\mathcal{L}_{\text{diffusion}},
  \label{eq:total_loss}
\end{equation}
where $\lambda_{\text{depth}} = 0.05$, $\lambda_{\text{latent}} = 0.1$, and $\lambda_{\text{diffusion}} = 1.0$ in our experiments.
The first term $\mathcal{L}_{\text{depth}}$ supervises the rendered depth map $\mathbf{D}^\text{nov}$ to train LRM. We use the scale-free loss:
\begin{equation}
  \mathcal{L}_{\text{depth}} = \sum_{j\in\mathcal{S}^\text{nov}} \left\| \pi(1/{\mathbf{D}^\text{nov}_j}) - \pi(1/\hat{\mathbf{D}}^\text{nov}_j)\right\|^2,
  \label{eq:depth_loss}
\end{equation}
where $\mathcal{S}^\text{nov}$ denotes an evenly sampled subset of novel views. $|\mathcal{S}^\text{nov}|=3$ in our experiments for efficiency. $\hat{\mathbf{D}}^\text{nov}_j$ is predicted by Depth Anything V2. $\pi(\cdot)$ normalizes the inverse depth to $[0,1]$, defined as:
\begin{equation}
  \pi(1/\mathbf{D}) = \frac{1/\mathbf{D} - \min(1/\mathbf{D})}{\max(1/\mathbf{D}) - \min(1/\mathbf{D})}.
\end{equation}
The second term $\mathcal{L}_{\text{latent}}$ supervises the latent $\mathbf{z}^\text{spat}$ to train both LRM and causal compression network:
\begin{equation}
  \mathcal{L}_{\text{latent}} = \left\| \mathbf{z}^\text{spat} - \mathcal{E}(\mathbf{x}) \right\|^2,
  \label{eq:latent_loss}
\end{equation}
where $\mathcal{E}$ denotes the pretrained VAE encoder, and $\textbf{x}$ is the groundtruth novel view images.
The third term $\mathcal{L}_{\text{diffusion}}$ is the traditional diffusion loss:
\begin{equation}
  \mathcal{L}_{\text{diffusion}} = \mathbb{E}_{\mathbf{x}, \, t, \, \mathbf{T}, \, \mathbf{z}^\text{st}, \, \epsilon} \left\|  \epsilon_{\theta} \left( \mathbf{z}_t, t,\mathbf{T}, \mathbf{z}^\text{st} \right) - \epsilon \right\|^2,
  \label{eq:diffusion_loss}
\end{equation}
where $\epsilon_{\theta}$ is the denoising network with model parameters $\theta$, $\mathbf{z}_t$ is the noisy latent corrupted by known noise $\epsilon$, and $t$ denotes the diffusion timestep.
}

\subsection{Downstream Tasks}
\label{sec:method-task}
\noindent\textbf{Sparse view interpolation} is a key step in sparse view reconstruction, \rev{as demonstrated} by concurrent works~\cite{DBLP:journals/corr/abs-2408-16767,DBLP:journals/corr/abs-2409-02048} closely related to ours. Given the start and end frames, it generates the intermediate images.
\rev{In the framework of StarGen, we distinguish two cases based on overlap between the two input images. In cases where the two input images have large overlapping regions, they serve as both spatial and temporal conditions, which are}
fed into the spatiotemporal-conditioned video generation model to generate an interpolated video.
For long-range scenes where the start and end frames
\rev{share minimal or even no common region,}
we propose a two-pass approach.
In the first pass, the process is similar to the previous approach, but the pose difference between adjacent generated images is larger \rev{compared to the first case}, resulting in a set of sparsely sampled images.
In the second pass, each pair of adjacent images from the first pass is treated as the start and end frames of a clip. All clips are generated as the first case.

\noindent\textbf{Perpetual view generation} is the task of generating novel views of a scene from a single image while allowing pose control~\cite{DBLP:conf/iccv/LiuM0SJK21,DBLP:journals/corr/abs-2409-02048}.
In the framework of StarGen, the input image serves as both the spatial and temporal condition for the first generated clip. The remaining clips are then generated using the proposed spatiotemporal autoregression.
\rev{Compared to the task of sparse view interpolation, where both the start and end frames provide constraints, perpetual view generation only has a constraint on the first frame. 
As a result, it is sensitive to error accumulation and places high demands on the scalability of the generation model.
}

\noindent\textbf{Layout-conditioned city generation} is the task of generating images given city layout and observation poses~\cite{DBLP:conf/cvpr/Xie0H024,DBLP:conf/siggraph/Deng0LGSW24}. 
\rev{
First, we render the city layout into depth and semantic videos based on the observation poses.
We then train two separate ControlNets for the depth and semantic maps, similar to the approach in~\cite{DBLP:journals/corr/abs-2401-05252}.
These two ControlNets are combined to generate the first clip.
The remaining clips are then generated using the proposed spatiotemporal autoregression.
The spatiotemporal condition can be effectively combined with the depth and semantic conditions, thanks to the combination capabilities of ControlNets and the flexibility of our proposed framework.
}

\section{Experiments}

We first describe the experiment setup in~\cref{sec_setup}, followed by qualitative and quantitative evaluations for the three downstream tasks in~\cref{sec_svi,sec_pvg,sec_lccg}, respectively. Finally, we conduct ablation studies in~\cref{sec_ablation}.


\begin{table}[t]
    \centering
    \scalebox{0.85}{
    \begin{tabular}{p{2.4cm}|p{0.75cm}p{0.75cm}p{0.87cm}|p{0.75cm}p{0.75cm}p{0.75cm}}
    \toprule
         Dataset&\multicolumn{3}{c|}{RealEstate-10K} &\multicolumn{3}{c}{ACID}\\
         \midrule
          Method & PSNR$\uparrow$ & SSIM$\uparrow$ & LPIPS$\downarrow$ & PSNR$\uparrow$ & SSIM$\uparrow$ & LPIPS$\downarrow$\\
         \midrule
          pixelNeRF\textsuperscript{*}~\cite{DBLP:conf/cvpr/YuYTK21}&20.43 & 0.589 & 0.550 & 20.97 & 0.547 & 0.533\\
         GPNR\textsuperscript{*}~\cite{suhail2022generalizable}& 24.11 & 0.793 & 0.255 & 25.28 & 0.764 & 0.332\\
         AttnRend\textsuperscript{*}~\cite{DBLP:conf/cvpr/DuSTS23} & 24.78 & 0.820 & 0.213 & 26.88 & 0.799 & 0.218\\
         MuRF\textsuperscript{*}~\cite{DBLP:conf/cvpr/XuCCSZP0024}& 26.10 & 0.858 & 0.143 & 28.09 & 0.841 & 0.155\\
         pixelSplat\textsuperscript{†}~\cite{DBLP:conf/cvpr/CharatanLTS24} & 25.89 & 0.858 & 0.142 & 28.14 & 0.839 & 0.150\\
         MVSplat\textsuperscript{†}~\cite{DBLP:journals/corr/abs-2403-14627}& 26.39 & 0.839 & 0.128 & 28.25 & 0.843 & 0.144 \\
         GS-LRM\textsuperscript{†}~\cite{DBLP:conf/eccv/ZhangBTXZSX24}& 28.10 & 0.892 & \underline{0.114} & - & - & - \\
         DepthSplat\textsuperscript{†}~\cite{xu2024depthsplat} & 27.44 & 0.887 & 0.119 & - & - & - \\
         \midrule
         ReconX\textsuperscript{†}~\cite{DBLP:journals/corr/abs-2408-16767}& \underline{28.31} & \textbf{0.912} & \textbf{0.088} & \underline{28.84} & \textbf{0.891} & \textbf{0.101} \\
         ViewCrafter\textsuperscript{\#}~\cite{DBLP:journals/corr/abs-2409-02048} & 24.23 & 0.790 & 0.217 & 23.48 & 0.660 & 0.299 \\
         StarGen (ours)\textsuperscript{\#}&  \textbf{28.49} & \underline{0.894} & \textbf{0.088} & \textbf{29.69} & \underline{0.876} & \underline{0.116} \\
         \bottomrule
    \end{tabular}
    }
    \caption{Quantitative comparison of sparse view interpolation. The upper part shows pure reconstruction-based methods, and the lower part shows the combined reconstruction and generation-based methods. Superscript \textsuperscript{*} indicates results from MVSplat~\cite{DBLP:journals/corr/abs-2403-14627}, \textsuperscript{†} refers to results from their original papers, and \textsuperscript{\#} denotes results run by ourselves.}
    \vspace{1em}
    \label{tab:nvs}
\end{table}

\begin{table}[t]

    \centering
    \scalebox{0.82}{
    \begin{tabular}{p{2.75cm}|p{0.75cm}p{0.75cm}p{0.87cm}|p{0.75cm}p{0.75cm}p{0.75cm}}
    \toprule
         Dataset & \multicolumn{3}{c|}{RealEstate-10K}& \multicolumn{3}{c}{Tanks-and-Temples}\\
         \midrule
          Method&  PSNR$\uparrow$ & SSIM$\uparrow$ & LPIPS$\downarrow$ &  PSNR$\uparrow$ & SSIM$\uparrow$ & LPIPS$\downarrow$ \\
         \midrule
         InfNat0 \cite{DBLP:conf/eccv/LiWSK22} & 12.29 & 0.435 & 0.632 & 10.78 & 0.251 & 0.699 \\
         LucidDreamer \cite{DBLP:conf/cvpr/LiangYLLXC24} & 22.27 & \underline{0.766} & 0.204 & 16.13 & 0.482 & 0.385  \\
         MotionCtrl \cite{DBLP:conf/siggraph/WangYWLCXLS24} & 15.86 & 0.520 & 0.431 & 13.02 & 0.321 & 0.584 \\
         \midrule
         ViewCrafter \cite{DBLP:journals/corr/abs-2409-02048} & \underline{22.60} & 0.754 & \underline{0.195} & \underline{17.25} & \underline{0.489} & \underline{0.341}\\
         StarGen (ours) & \textbf{23.52} & \textbf{0.792} & \textbf{0.162} &\textbf{19.52}& \textbf{0.552} & \textbf{0.332} \\
         \bottomrule
    \end{tabular}
    }
    \caption{
    Quantitative comparison of perpetual view generation. Both ViewCrafter and our method are trained on RealEstate-10K and not on the Tanks-and-Temples dataset, demonstrating the generalization capability of both methods.
    }
    \vspace{1em}
    \label{tab:Zero_Shot}
\end{table}

\begin{figure*}[htbp]
    \centering
    \begin{tikzpicture}[scale=1, every node/.style={transform shape}]

    \def\boxwidth{1.7 cm} 
    \def\boxheight{1.7 cm} 
    \def\horizontaldistance{0.1cm} 
    \def\verticaldistance{0.1cm} 
    \def\largedistance{0.25cm} 
    \def\smalldistance{0.02cm} 

    \node[font=\small] at (1*\boxwidth, -6.0*\verticaldistance) {Input};
    \node[font=\small] at (2.5*\boxwidth + 1*\largedistance, -6.0*\verticaldistance) {Groundtruth};
    \node[font=\small] at (4.5*\boxwidth + 2*\largedistance, -6.0*\verticaldistance) {StarGen (ours)};
    \node[font=\small] at (6.5*\boxwidth + 3*\largedistance, -6.0*\verticaldistance) {ViewCrafter~\cite{DBLP:journals/corr/abs-2409-02048}};
    \node[font=\small] at (8.5*\boxwidth + 4*\largedistance, -6.0*\verticaldistance) {MVSplat~\cite{DBLP:journals/corr/abs-2403-14627}};

    \newcommand{\imagepath}[1]{
    \ifcase#1 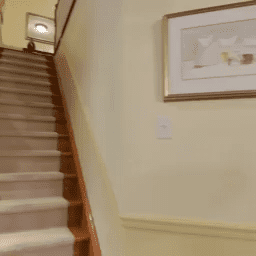\or
              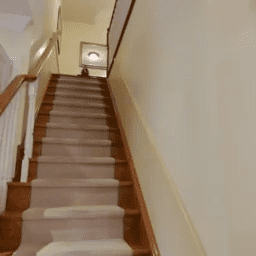\or
              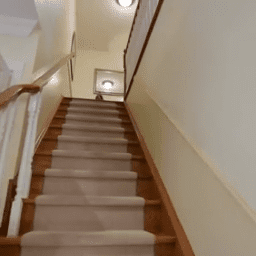\or
              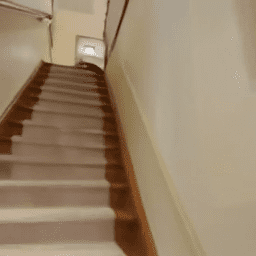\or
              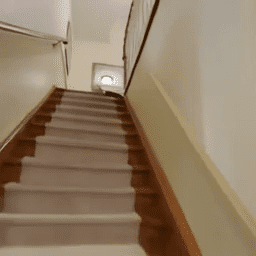\or
              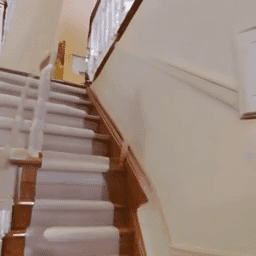\or
              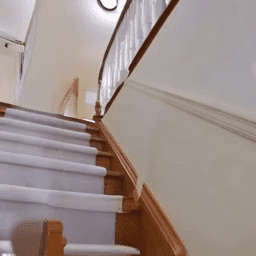\or
              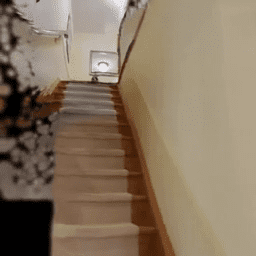\or
              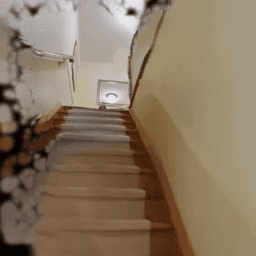\or
              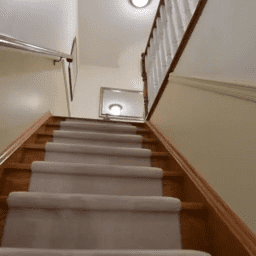\or
              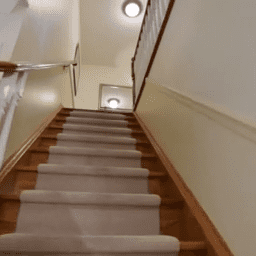\or
              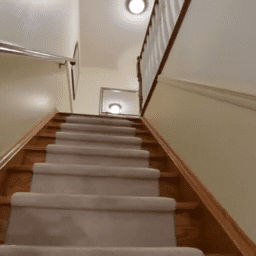\or
              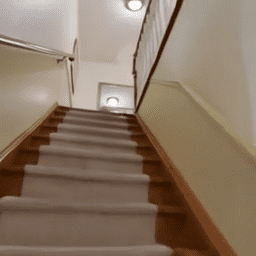\or
              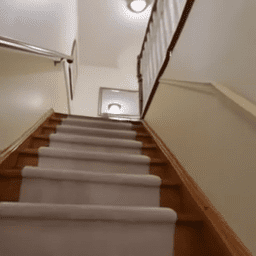\or
              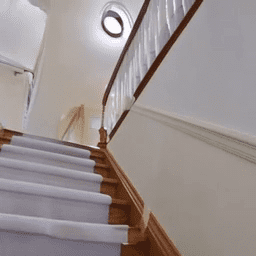\or
              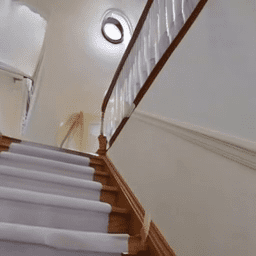\or
              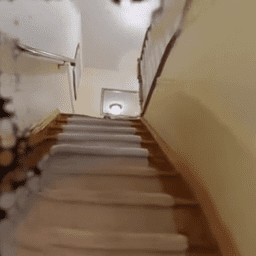\or
              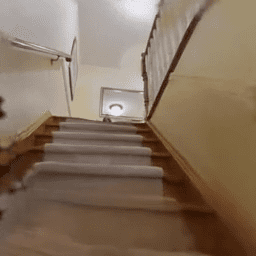\or
              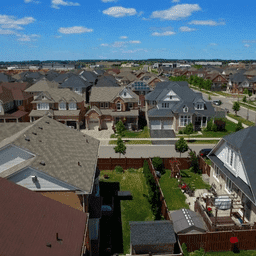\or
              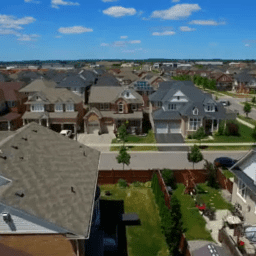\or
              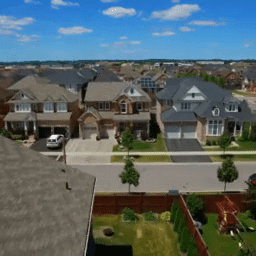\or
              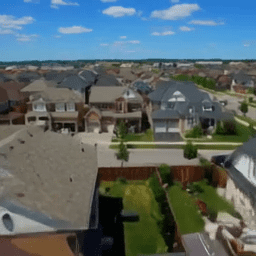\or
              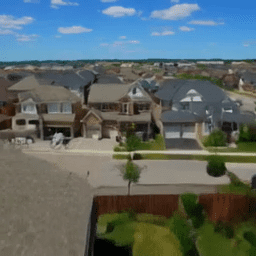\or
              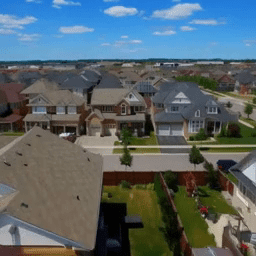\or
              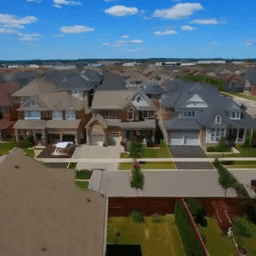\or
              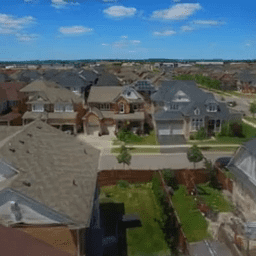\or
              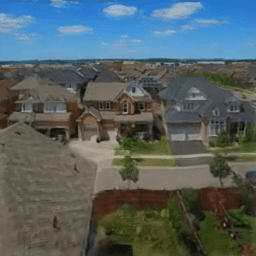\or
              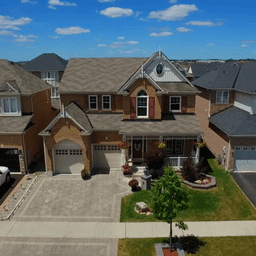\or
              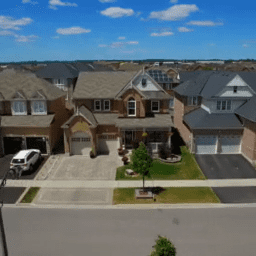\or
              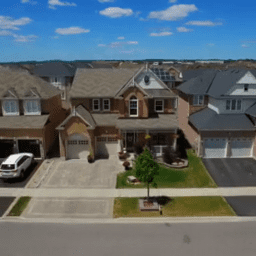\or
              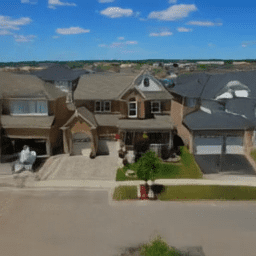\or
              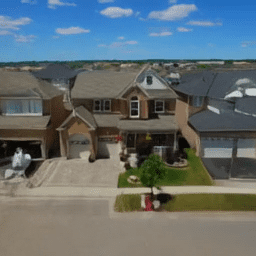\or
              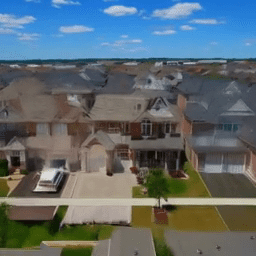\or
              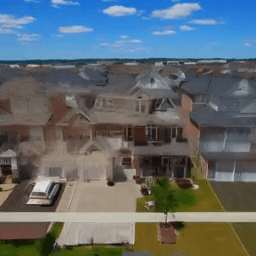\or
              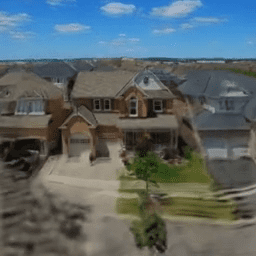\or
              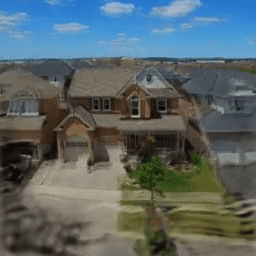\or
              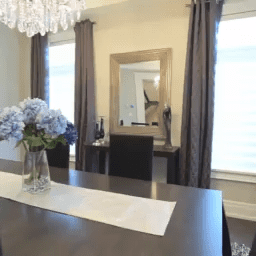\or
              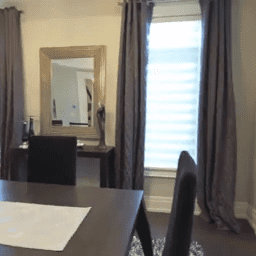\or
              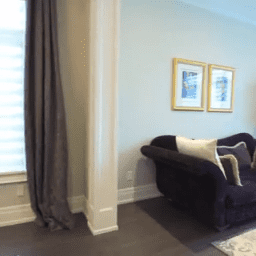\or
              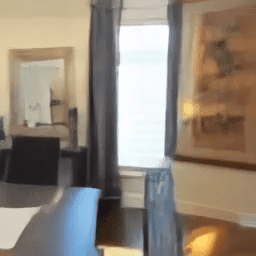\or
              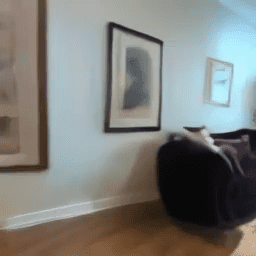\or
              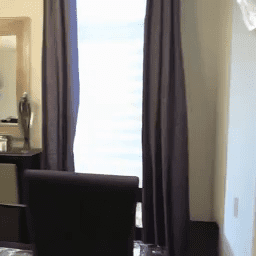\or
              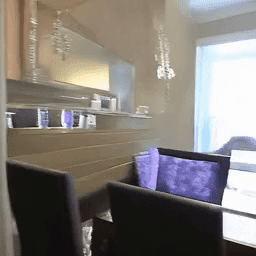\or
              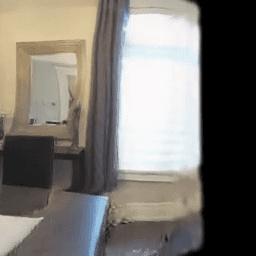\or
              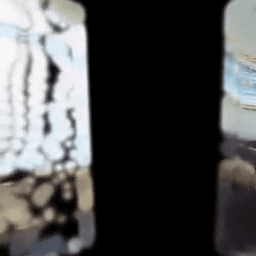\or
              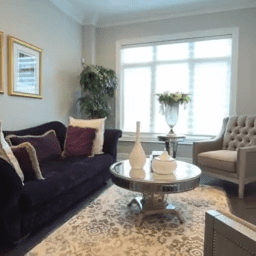\or
              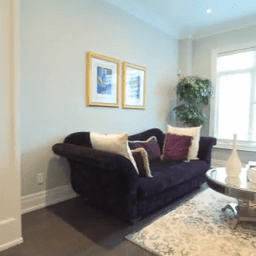\or
              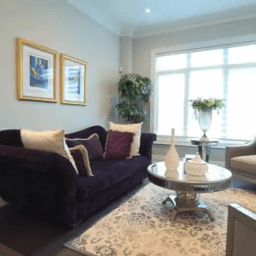\or
              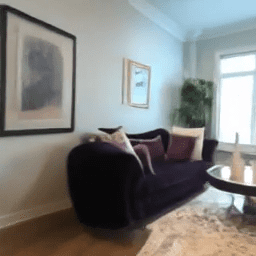\or
              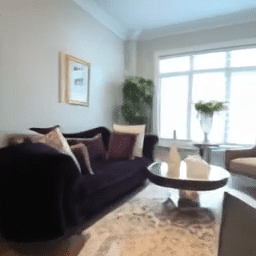\or
              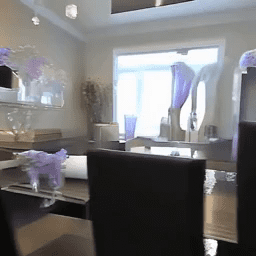\or
              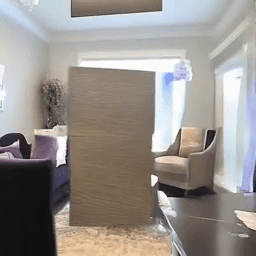\or
              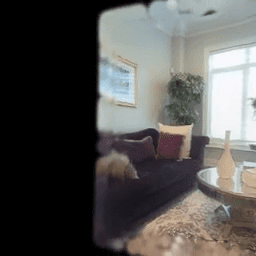\or
              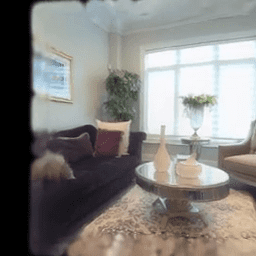\or
    \fi
    }
    \foreach \row in {1, 2, 3, 4, 5, 6} {
        \pgfmathsetmacro{\ypos}{-1*\row*\boxheight - (\row-1)*\verticaldistance}
        \foreach \col in {1, 2, 3, 4, 5, 6, 7, 8, 9} {
        \pgfmathtruncatemacro{\imgindex}{(\row - 1) * 9 + \col - 1}
            \node[minimum width=\boxwidth, minimum height=\boxheight] 
                at ({\col * \boxwidth + floor((\col) / 2)*\largedistance + floor((\col - 1) / 2) * \smalldistance},{-1*\row*\boxheight - floor((\row - 1)/2)*\largedistance - floor((\row) / 2) * \smalldistance}) 
                {
                \includegraphics[width=\boxwidth, height=\boxheight]{\imagepath{\imgindex}} 
            };
        }
    }
    \draw[line width=1.0pt, black] 
        ({0.5*\boxwidth - 0.2*\largedistance}, {-0.5*\boxheight + 0.5*\verticaldistance}) 
        rectangle 
        ({1.5*\boxwidth + 0.2*\largedistance}, {-2.5*\boxheight - 0.8*\verticaldistance});

    \draw[line width=1.0pt, black] 
        ({0.5*\boxwidth - 0.2*\largedistance}, {-2.5*\boxheight - 2.1*\verticaldistance}) 
        rectangle 
        ({1.5*\boxwidth + 0.2*\largedistance}, {-4.5*\boxheight - 3.5*\verticaldistance});

    \draw[line width=1.0pt, black] 
        ({0.5*\boxwidth - 0.2*\largedistance}, {-4.5*\boxheight - 4.9*\verticaldistance}) 
        rectangle 
        ({1.5*\boxwidth + 0.2*\largedistance}, {-6.5*\boxheight - 6.2*\verticaldistance});

    \draw[line width=1.0pt, black] 
        ({1.5*\boxwidth + 0.8*\largedistance}, {-0.5*\boxheight + 0.5*\verticaldistance}) 
        rectangle 
        ({3.5*\boxwidth + 1.3*\largedistance}, {-2.5*\boxheight - 0.8*\verticaldistance});

    \draw[line width=1.0pt, black] 
        ({1.5*\boxwidth + 0.8*\largedistance}, {-2.5*\boxheight - 2.1*\verticaldistance}) 
        rectangle 
        ({3.5*\boxwidth + 1.3*\largedistance}, {-4.5*\boxheight - 3.5*\verticaldistance});

    \draw[line width=1.0pt, black] 
        ({1.5*\boxwidth + 0.8*\largedistance}, {-4.5*\boxheight - 4.9*\verticaldistance}) 
        rectangle 
        ({3.5*\boxwidth + 1.3*\largedistance}, {-6.5*\boxheight - 6.2*\verticaldistance});
        
    \draw[line width=1.0pt, black] 
        ({3.5*\boxwidth + 1.8*\largedistance}, {-0.5*\boxheight + 0.5*\verticaldistance}) 
        rectangle 
        ({5.5*\boxwidth + 2.4*\largedistance}, {-2.5*\boxheight - 0.8*\verticaldistance});

    \draw[line width=1.0pt, black] 
        ({3.5*\boxwidth + 1.8*\largedistance}, {-2.5*\boxheight - 2.1*\verticaldistance}) 
        rectangle 
        ({5.5*\boxwidth + 2.4*\largedistance}, {-4.5*\boxheight - 3.5*\verticaldistance});

    \draw[line width=1.0pt, black] 
        ({3.5*\boxwidth + 1.8*\largedistance}, {-4.5*\boxheight - 4.9*\verticaldistance}) 
        rectangle 
        ({5.5*\boxwidth + 2.4*\largedistance}, {-6.5*\boxheight - 6.2*\verticaldistance});
        
    \draw[line width=1.0pt, black] 
        ({5.5*\boxwidth + 2.95*\largedistance}, {-0.5*\boxheight + 0.5*\verticaldistance}) 
        rectangle 
        ({7.5*\boxwidth + 3.5*\largedistance}, {-2.5*\boxheight - 0.8*\verticaldistance});

    \draw[line width=1.0pt, black] 
        ({5.5*\boxwidth + 2.95*\largedistance}, {-2.5*\boxheight - 2.1*\verticaldistance}) 
        rectangle 
        ({7.5*\boxwidth + 3.5*\largedistance}, {-4.5*\boxheight - 3.5*\verticaldistance});
        
    \draw[line width=1.0pt, black] 
        ({5.5*\boxwidth + 2.95*\largedistance}, {-4.5*\boxheight - 4.9*\verticaldistance}) 
        rectangle 
        ({7.5*\boxwidth + 3.5*\largedistance}, {-6.5*\boxheight - 6.2*\verticaldistance});
        
    \draw[line width=1.0pt, black] 
        ({7.5*\boxwidth + 4*\largedistance}, {-0.5*\boxheight + 0.5*\verticaldistance}) 
        rectangle 
        ({9.5*\boxwidth + 4.52*\largedistance}, {-2.5*\boxheight - 0.8*\verticaldistance});
        
    \draw[line width=1.0pt, black] 
        ({7.5*\boxwidth + 4*\largedistance}, {-2.5*\boxheight - 2.1*\verticaldistance}) 
        rectangle 
        ({9.5*\boxwidth + 4.52*\largedistance}, {-4.5*\boxheight - 3.5*\verticaldistance});
    
    \draw[line width=1.0pt, black] 
        ({7.5*\boxwidth + 4*\largedistance}, {-4.5*\boxheight - 4.9*\verticaldistance}) 
        rectangle 
        ({9.5*\boxwidth + 4.52*\largedistance}, {-6.5*\boxheight - 6.2*\verticaldistance});
    \node[anchor=east, rotate=90, font=\small] at (0.35*\boxwidth, -0.6*\boxheight - 0.0*\verticaldistance) { (a) ca38b62977d9bdaa7}; 
    \node[anchor=east, rotate=90, font=\small] at (0.35*\boxwidth, -2.8*\boxheight - 0.0*\verticaldistance) { (b) 36bc6918e9fc837c}; 
    \node[anchor=east, rotate=90, font=\small] at (0.35*\boxwidth, -4.6*\boxheight - 5*\verticaldistance) { (c) d235e2853aa439259 }; 
    \end{tikzpicture}
    \caption{
    Qualitative comparison of sparse view interpolation on the RealEstate-10K~\cite{DBLP:journals/tog/ZhouTFFS18} test dataset under challenging scenario where the two input images have minimal or no overlap.
    In these situations, our method demonstrates better performance compared to other methods.
    We encourage readers to watch our supplementary video to better appreciate the differences.}
    \label{fig:large_angle}
    \vspace{-1em}
\end{figure*}

\begin{figure}[htbp]
    \centering
    \includegraphics[width=1.0\columnwidth]{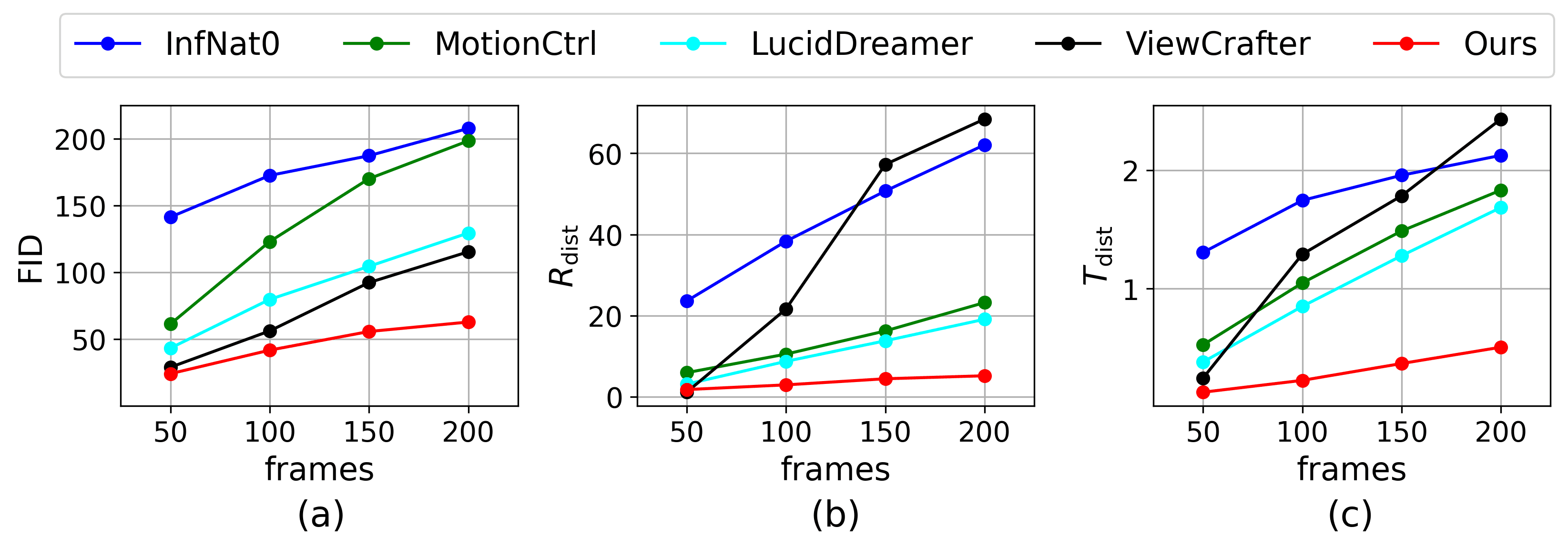}
    \caption{
    Scalability comparison of perpetual view generation on long-range videos on the RealEstate-10K~\cite{DBLP:journals/tog/ZhouTFFS18} test dataset.
    For a fair FID comparison across different desired numbers of frames, for each desired frame number $N$, we generate 5K$/N$ results for each method.
    Our method significantly outperforms existing methods in terms of both fidelity (a) and pose accuracy (b)(c).
    }
  \label{fig:fid_long}
\end{figure}

\begin{figure*}[htbp]
    \centering
    \begin{tikzpicture}[scale=1, every node/.style={transform shape}]

    \def\boxwidth{1.7 cm} 
    \def\boxheight{1.7 cm} 
    \def\horizontaldistance{0.05 cm} 
    \def\verticaldistance{0.05 cm} 
    \def\largedistance{0.15cm} 
    \def\smalldistance{0.05cm} 

    \node[font=\small] at (1*\boxwidth + 0.0*\horizontaldistance, -12*\verticaldistance) {Input};
    \node[font=\small] at (3.5*\boxwidth + 1.0*\horizontaldistance + 1.0 * \largedistance , -12*\verticaldistance) {StarGen (ours)};
    \node[font=\small] at (7.5*\boxwidth + 3.0*\horizontaldistance + 2.0 * \largedistance, -12*\verticaldistance) {ViewCrafter~\cite{DBLP:journals/corr/abs-2409-02048}};


    \node[anchor=east, rotate=90, font=\tiny] at (0.4*\boxwidth, -0.5*\boxheight + 0*\verticaldistance) {(a)1edc6b95e84127b6}; 
    \node[anchor=east, rotate=90, font=\tiny] at (0.4*\boxwidth, -1.5*\boxheight - 1*\verticaldistance) {(b)22a0db80d91128e4}; 
    \node[anchor=east, rotate=90, font=\tiny] at (0.4*\boxwidth, -2.5*\boxheight - 2*\verticaldistance) {(c)166818269e4e2568}; 

    \newcommand{\imagepath}[1]{
    \ifcase#1 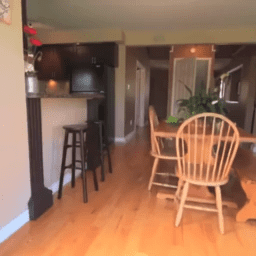\or
              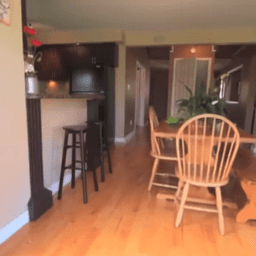\or
              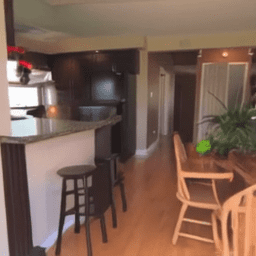\or
              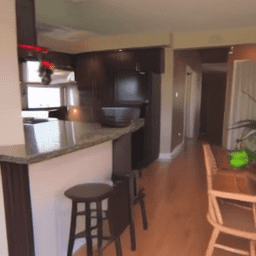\or
              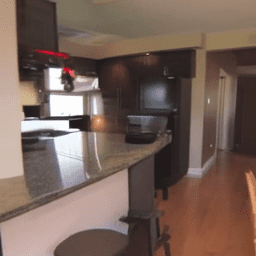\or
              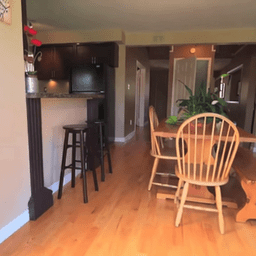\or
              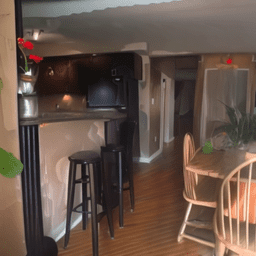\or
              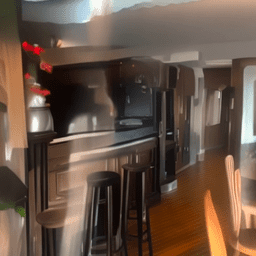\or
              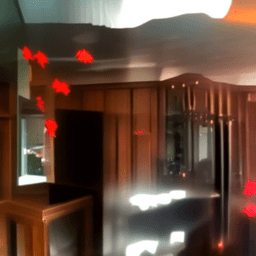\or
              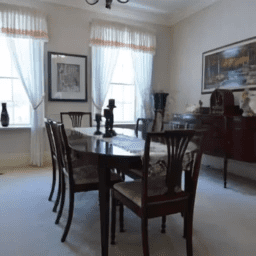\or
              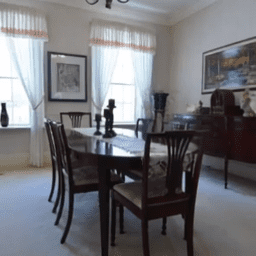\or
              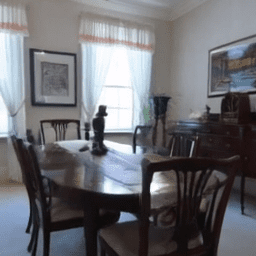\or
              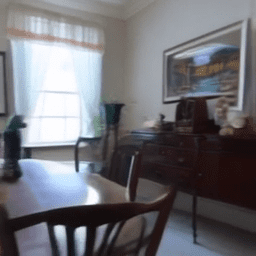\or
              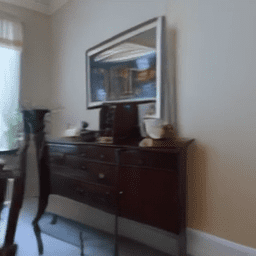\or
              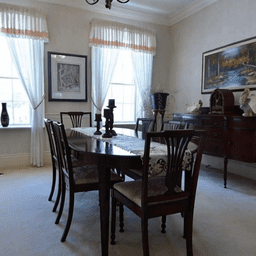\or
              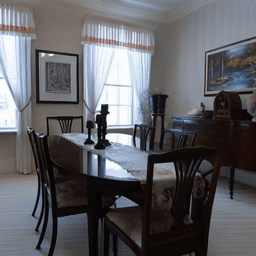\or
              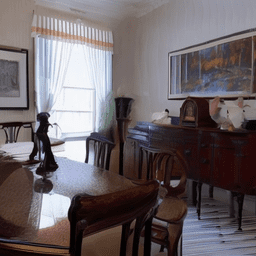\or
              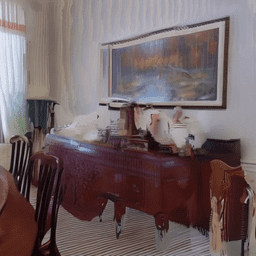\or
              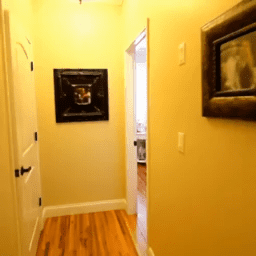\or
              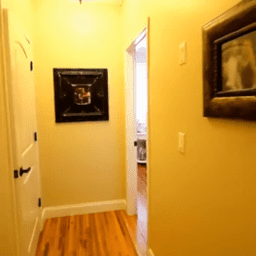\or
              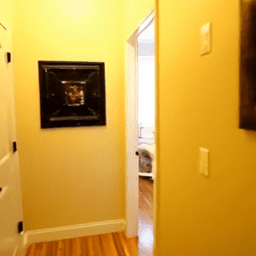\or
              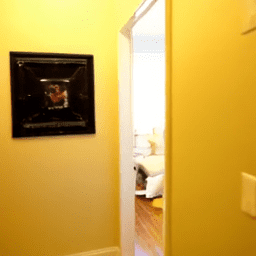\or
              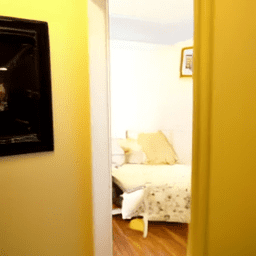\or
              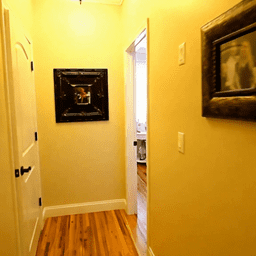\or
              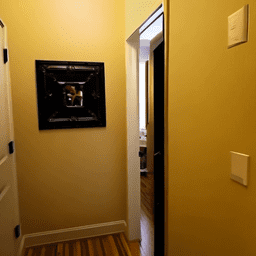\or
              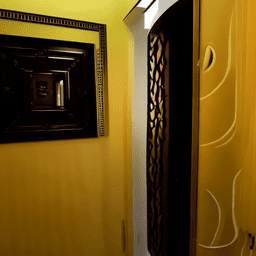\or
              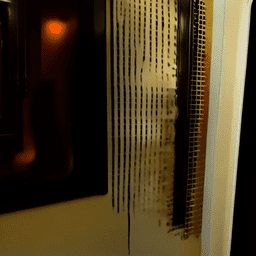\or
    \fi
    }
    
    \foreach \row in {1, 2, 3} {
        \foreach \col in {1, 2, 3, 4, 5, 6, 7,8,9} {
        \pgfmathtruncatemacro{\imgindex}{(\row - 1) * 9 + \col - 1}
            \node[minimum width=\boxwidth, minimum height=\boxheight] 
                at ({\col * \boxwidth + floor((\col+2) / 4)*\largedistance + \col * \smalldistance},{-1*\row*\boxheight - (\row - 1)*\verticaldistance}) {
                
                \includegraphics[width=\boxwidth, height=\boxheight]{\imagepath{\imgindex}} 
                };
        }
    }
    \draw[line width=1.0pt, black] 
        ({1*\boxwidth - 0.5*\boxwidth + 0.25 * \largedistance}, {-0.5*\boxheight + 0.25 *\verticaldistance}) 
        rectangle ({1*\boxwidth + 0.25 * \largedistance+ 0.5*\boxwidth}, 
        {-3.5*\boxheight - 2.25 *\verticaldistance});
    \draw[line width=1.0pt, black] 
        ({1.5*\boxwidth + 1.25 * \largedistance + 1.0 *\horizontaldistance}, {-0.5*\boxheight + 0.25*\verticaldistance}) 
        rectangle 
        ({5.5*\boxwidth + 2.25*\largedistance + 1.0 * \horizontaldistance}, {-3.5*\boxheight - 2.25*\verticaldistance});
    \draw[line width=1.0pt, black] 
        ({5.5*\boxwidth + 3.25*\largedistance + 2.0 *\horizontaldistance}, {-0.5*\boxheight + 0.25*\verticaldistance}) 
        rectangle 
        ({9.5*\boxwidth + 3.25*\largedistance + 5.5 * \horizontaldistance}, {-3.5*\boxheight - 2.25*\verticaldistance});

    \end{tikzpicture}
    \caption{
    Qualitative comparison of perpetual view generation on long-range videos on the RealEstate-10K~\cite{DBLP:journals/tog/ZhouTFFS18} test dataset. 
    While ViewCrafter exhibits significant degradation as the generated video becomes longer, our method is able to generate reasonable content throughout the entire sequence. We encourage readers to watch our supplementary video to better appreciate the differences.
    }
    \label{fig:ar}
    \vspace{-1em}
\end{figure*}

\begin{table}[t]

    \centering
    \scalebox{0.8}{
    \begin{tabular}{l|p{0.75cm}p{0.75cm}p{0.75cm}|p{0.75cm}p{0.75cm}p{0.75cm}}
    \toprule
         Dataset & \multicolumn{3}{c|}{RealEstate-10K}& \multicolumn{3}{c}{ACID}\\
         \midrule
         Setup & FID$\downarrow$ & $R_\text{dist}$$\downarrow$ & $T_\text{dist}$$\downarrow$  & FID$\downarrow$ & $R_\text{dist}$$\downarrow$ & $T_\text{dist}$$\downarrow$\\
         \midrule
          Ours & \textbf{41.72} & \textbf{2.088} & \textbf{0.453} & \textbf{40.73} & \textbf{3.542} & \textbf{0.618} \\
         w/o spatial cond. & \underline{43.21} & 11.70 & 1.643 & \underline{53.09} & 4.076 & 0.684 \\
         w/o temporal cond. & 54.12 & \underline{7.551} & \underline{1.314}  & 53.37 & \underline{3.593} & \underline{0.626} \\
         \bottomrule
    \end{tabular}
    }
    \caption{
    Ablation of the proposed spatiotemporal autoregression on the task of perpetual view generation with 100-frame videos.
    }
    \vspace{1em}
    \label{tab:Ablation_Study_AR}
\end{table}

\begin{table}

    \centering
    \scalebox{0.8}{
    \begin{tabular}{p{2.35cm}|p{0.75cm}p{0.75cm}p{0.87cm}|p{0.75cm}p{0.75cm}p{0.75cm}}
    \toprule
         Dataset & \multicolumn{3}{c|}{RealEstate-10K} & \multicolumn{3}{c}{ACID}\\
         \midrule
         Setup &  PSNR$\uparrow$ & SSIM$\uparrow$ & 
         LPIPS$\downarrow$ &  PSNR$\uparrow$ & SSIM$\uparrow$ & LPIPS$\downarrow$ \\
         \midrule
          Ours & \textbf{28.49} & \textbf{0.894} & \textbf{0.088} & \textbf{29.69} & \textbf{0.876} & \textbf{0.116} \\
         w/ DUSt3R \cite{DBLP:conf/cvpr/Wang0CCR24} & 26.94 & 0.863 & 0.123 & 26.74 & 0.823 & 0.157 \\
         w/o spatial cond. & 23.01 & 0.738 & 0.170 & 25.47 & 0.781 & 0.176 \\
         w/o depth input & 27.32 & 0.867 & 0.099 & 28.99 & 0.835 & 0.127 \\
         w/o depth loss & \underline{27.63} & \underline{0.887} & \underline{0.093} & \underline{29.12} & \underline{0.867} & \underline{0.121} \\
         fix LRM & 27.15 & 0.873 & 0.097 & 28.73 & 0.863 & 0.125 \\
         \bottomrule
    \end{tabular}
    }
    \caption{Ablation of the proposed spatiotemporal-conditioned video generation on the task of sparse view interpolation with single-clip videos.
    }
    \label{tab:Ablation_Study_LRM}
    \vspace{1em}
\end{table}

\begin{figure*}[htbp]
    \centering
    \begin{tikzpicture}[scale=1, every node/.style={transform shape}]

    \def\boxwidth{1.6 cm} 
    \def\boxheight{1.6 cm} 
    \def\horizontaldistance{0.05 cm} 
    \def\verticaldistance{0.05 cm} 
    \def\largedistance{0.15cm} 

    \node[font=\small] at (1.0*\boxwidth + 0.0 * \horizontaldistance, -10.0*\verticaldistance) {Layout};
    \node[font=\small] at (2.5*\boxwidth + 1.5 * \horizontaldistance, -10.0*\verticaldistance) {StarGen (ours)};
    \node[font=\small] at (4.5*\boxwidth + 3.5 * \horizontaldistance, -10.0*\verticaldistance) {CityDreamer~\cite{DBLP:conf/cvpr/Xie0H024}};
    \node[font=\small] at (6.0*\boxwidth + 5.0 * \horizontaldistance + \largedistance, -10.0*\verticaldistance) {Layout};
    \node[font=\small] at (7.5*\boxwidth + 6.5 * \horizontaldistance + \largedistance, -10.0*\verticaldistance) {StarGen (ours)};
    \node[font=\small] at (9.5*\boxwidth + 8.5 * \horizontaldistance + \largedistance, -10.0*\verticaldistance) {CityDreamer~\cite{DBLP:conf/cvpr/Xie0H024}};
    
    \newcommand{\imagepath}[1]{
    \ifcase#1 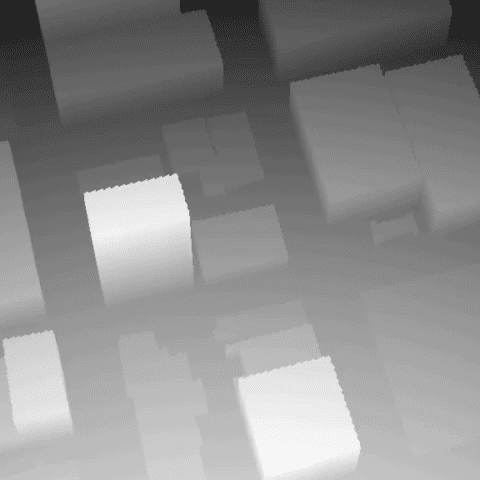\or
              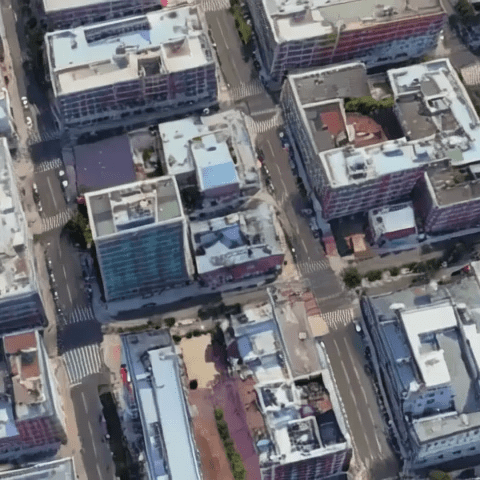\or
              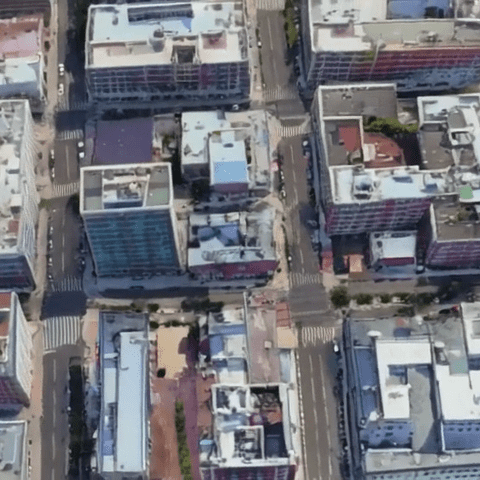\or
              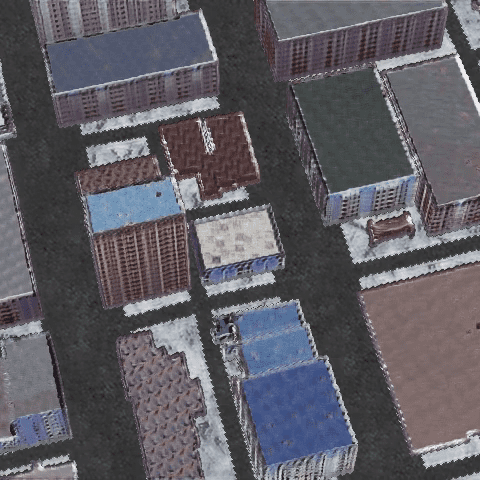\or
              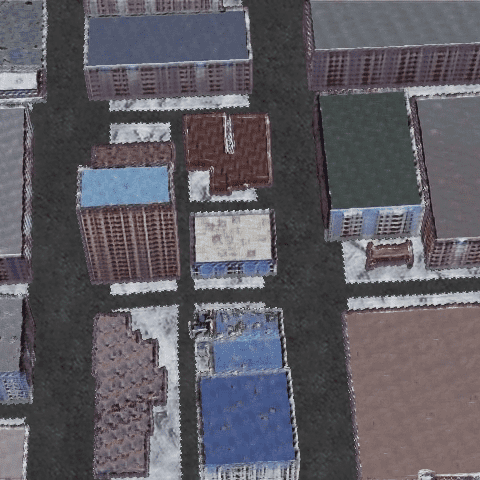\or
              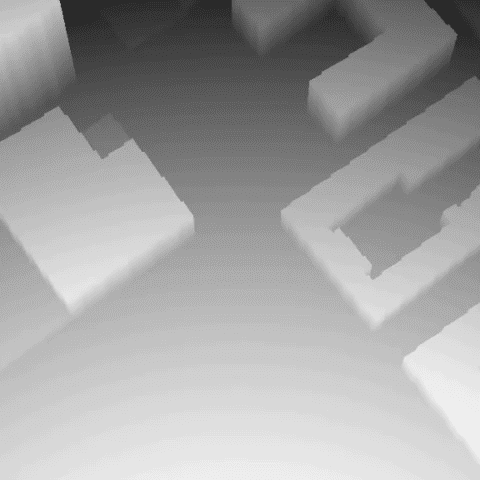\or
              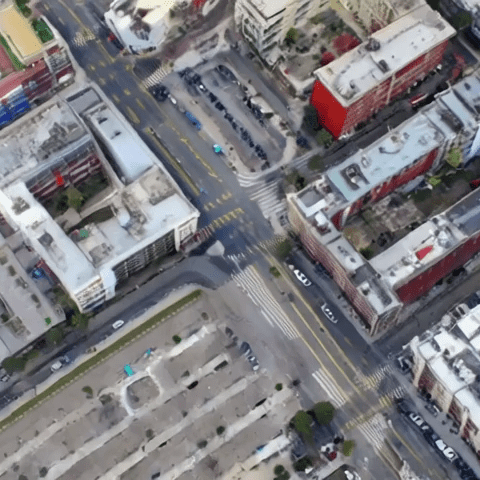\or
              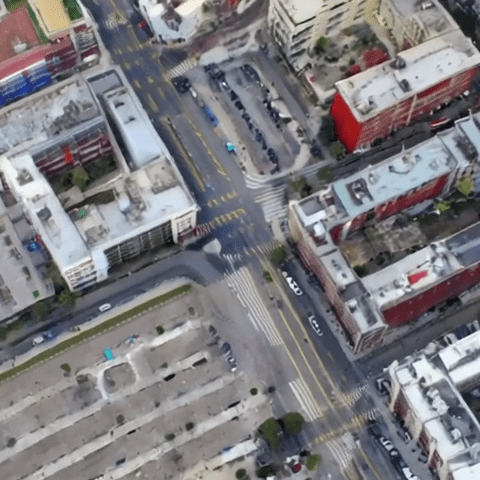\or
              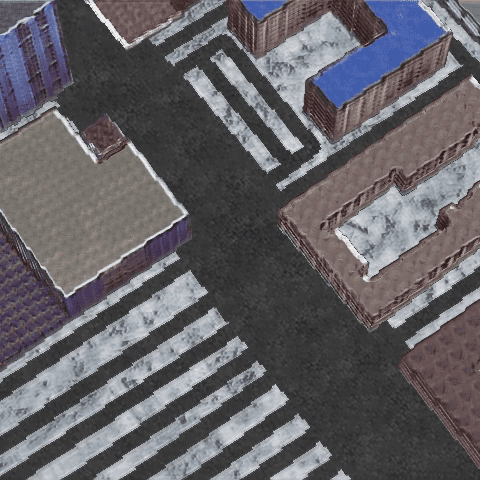\or
              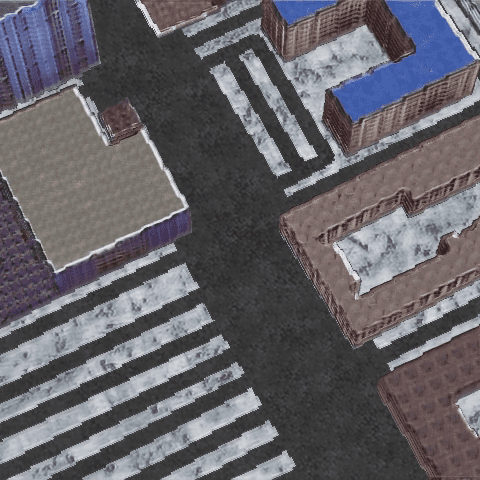\or
              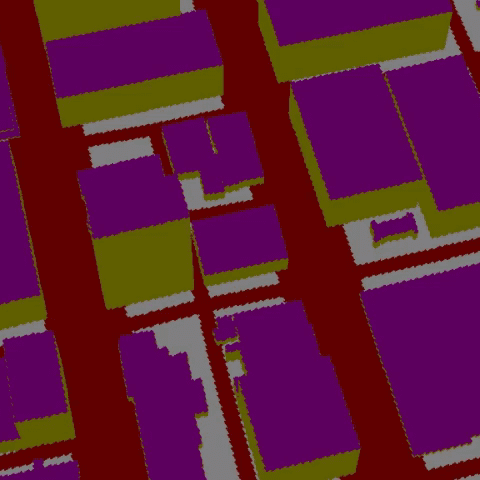\or
              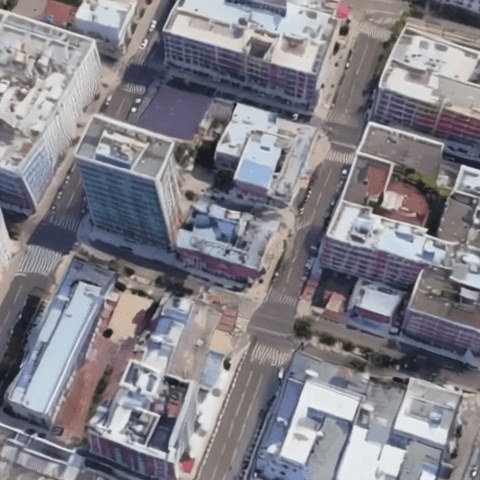\or
              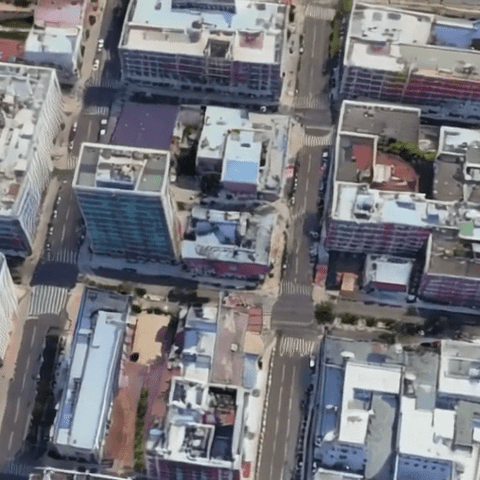\or
              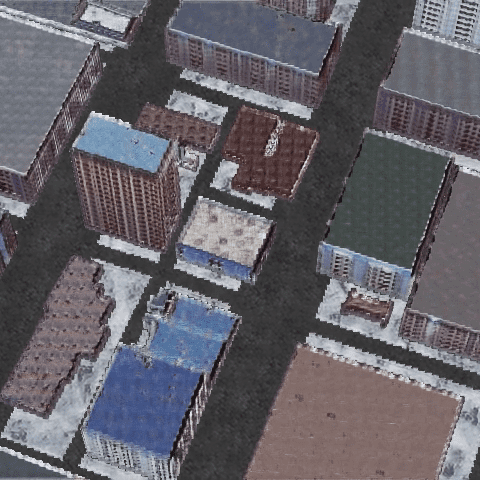\or
              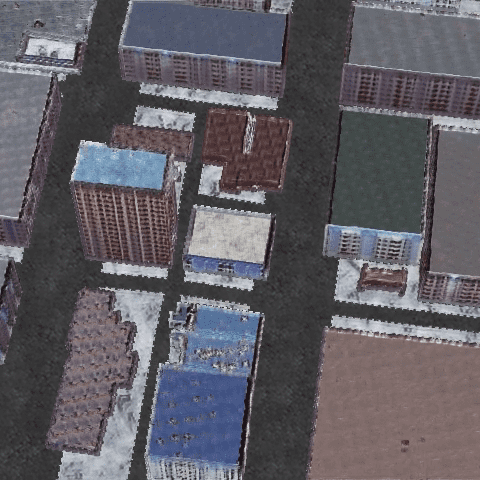\or
              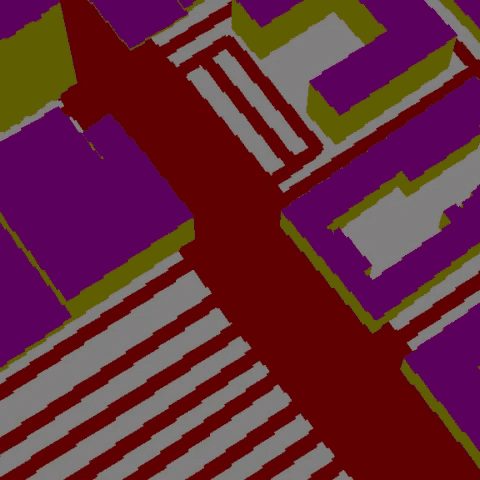\or
              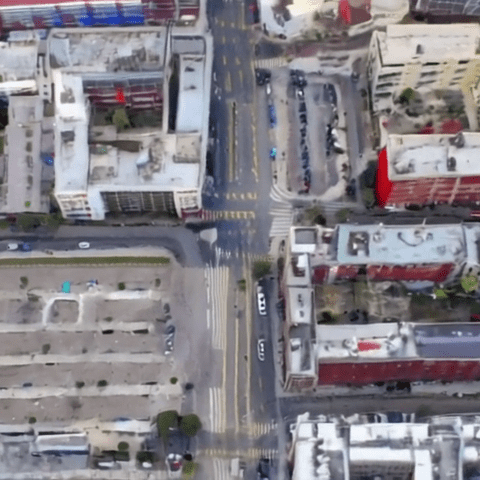\or
              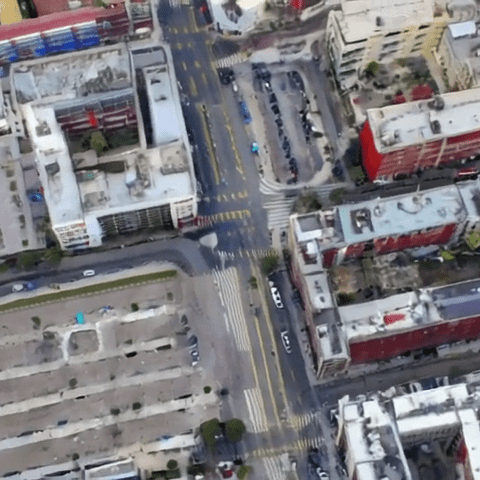\or
              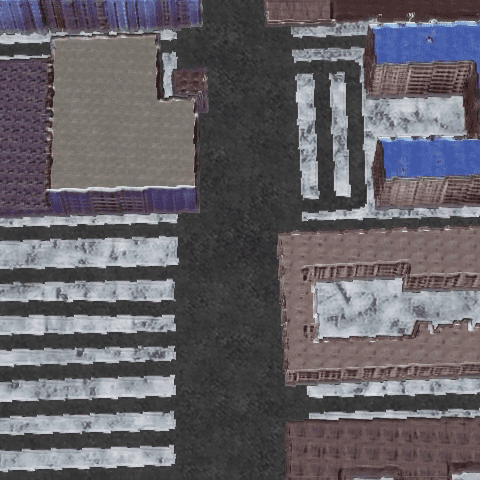\or
              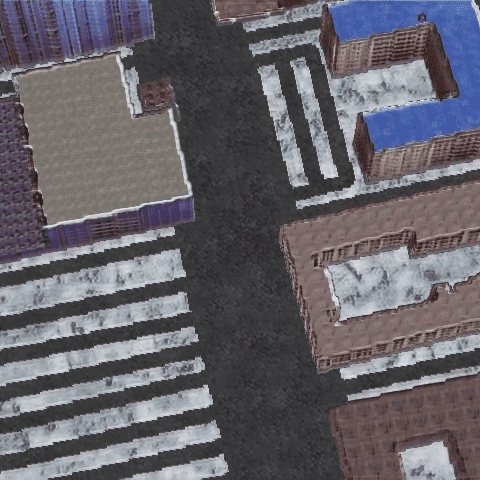\or              
    \fi
    }
    \foreach \row in {1,2} {
        \foreach \col in {1,2,3,4,5,6,7,8,9,10} {
        \pgfmathtruncatemacro{\imgindex}{(\row - 1) * 10 + \col - 1}
            \node[minimum width=\boxwidth, minimum height=\boxheight] 
                at ({\col * \boxwidth + (\col - 1) * \horizontaldistance
                + floor((\col) / 6)*\largedistance
                } ,{-1 * \row * \boxheight - (\row - 1) * \verticaldistance}) 
                {
                \includegraphics[width=\boxwidth, height=\boxheight]{\imagepath{\imgindex}} 
            };
        }
    }

    \draw[line width=1.0 pt, black] 
        ({1*\boxwidth - 0.5*\boxwidth - 1 * \horizontaldistance}, {-0.5*\boxheight + 1 *\verticaldistance}) 
        rectangle ({5 *\boxwidth + 5 *\horizontaldistance + 0.5*\boxwidth}, 
        {-2.5*\boxheight - 2 *\verticaldistance});
    \draw[line width=1.0 pt, black] 
        ({6 *\boxwidth - 0.5*\boxwidth + 5 * \horizontaldistance + \largedistance - 1 * \horizontaldistance}, {-0.5*\boxheight + 1 *\verticaldistance}) 
        rectangle ({10 *\boxwidth + 9 *\horizontaldistance + \largedistance + 0.5*\boxwidth + 1 * \horizontaldistance}, 
        {-2.5*\boxheight - 2 *\verticaldistance});

    \end{tikzpicture}
    \caption{
    Qualitative comparison of layout-conditioned city generation, where the depth and semantic maps are rendered from the OpenStreetMap dataset~\cite{osmdataset}.
    Both CityDreamer~\cite{DBLP:conf/cvpr/Xie0H024} and our method can effectively follow the desired layout, but our method achieves significantly better fidelity.
    We encourage readers to watch our supplementary video to better appreciate the differences.
    }
    \label{fig:layout}
    \vspace{-1em}
\end{figure*}


\subsection{Experiment Setup}
\label{sec_setup}
\textbf{Implementation Details.}
Our model is composed of a large reconstruction model (LRM), \rev{a causal compression network (CCN),} and a video diffusion model (VDM) with ControlNet. For the LRM, we use vision transformer architecture~\cite{DBLP:conf/iclr/DosovitskiyB0WZ21}, which consists of 12 layers with a hidden size of 768, an MLP size of 4096, and 12 attention heads, totaling 114M parameters.
\rev{
For the CCN, we implement a Conv3d layer with both input and output channels set to 16, a kernel size of (3, 3, 3), a stride of (1, 1, 1), a dilation of (1, 1, 1), and a padding of (0, 0, 0). 
For the VDM, we use the pretrained CogVideoX-2B~\cite{DBLP:journals/corr/abs-2408-06072} without any fine-tuning.
The ControlNet model contains 6 layers, which are the trainable copies of the first 6 layers of CogVideoX-2B.
}
\cam{
Note that different baseline methods employ different VDMs, which may introduce unfair comparisons. To mitigate this flaw, we provide an ablation study in the supplementary material (Sec. 6) to demonstrate that the improvement over the baseline methods is not solely attributable to the stronger VDM backbone.
}

\noindent\textbf{Training Details.} 
For the efficiency of training, the proposed model is trained at a resolution of 256$\times$256. The training process includes: 1) training the \rev{LRM+CCN} from scratch using consecutive frames with batch size 384, starting with 1K warm-up steps and continuing for a total of 40K steps; 2) training the \rev{LRM+CCN} using $1\sim3$ frame intervals, with batch size 384 for 20K steps; 3) joint training of the \rev{LRM+CCN} and ControlNet, also using $1\sim3$ frame intervals, with batch size 240 for 15K steps. For the task of layout-conditioned city generation, we additionally train two ControlNets for depth and semantic control, with batch size 128 over 20K steps. We also fine-tune our model by training the \rev{LRM+CCN} with batch size 16 for 11K steps, followed by separately training the ControlNet with a batch size of 80 for another 11K steps.

\noindent\textbf{Training Datasets.} 
The training data includes RealEstate-10K~\cite{DBLP:journals/tog/ZhouTFFS18}, ACID~\cite{DBLP:conf/iccv/LiuM0SJK21}, and DL3DV-10K~\cite{DBLP:conf/cvpr/LingSTZXWYGYLLS24}. 
We filter out short video clips, resulting in a final dataset of 66,859 videos.
For the task of layout-conditioned city generation, we additionally use the CityGen dataset from CityDreamer~\cite{DBLP:conf/cvpr/Xie0H024}, which consists of city layout data from OpenStreetMap~\cite{osmdataset} and renderings from Google Earth Studio~\cite{google_earth_stdio}.
The dataset includes 400 trajectories, each originally with 60 frames, which we interpolated to 600 frames.

\noindent\textbf{Metrics.} 
For short-range videos,
\rev{which are defined as videos with frames $N\leq 49$ (the maximum number of frames supported by CogVideoX in a single clip),}
we use PSNR, SSIM~\cite{DBLP:journals/tip/WangBSS04}, and LPIPS~\cite{DBLP:conf/cvpr/ZhangIESW18} metrics to assess the similarity between generated images and the groundtruth. 
\rev{For long-range perpetual view generation, which contains more than 49 frames and thus require multiple clips, most of the content is not visible in the first frame and is freely generated.}
We use FID~\cite{DBLP:conf/nips/HeuselRUNH17} to measure the distributional distance between the generated and groundtruth images. 
Each FID calculation is based on 5K images.
These metrics are calculated at a resolution of $256\times256$.
For models not originally designed for this resolution, the outputs are cropped to $256\times256$ before metric calculation. 
\rev{For long-range perpetual view generation,}
we also compute rotation distance $R_\text{dist}$ and translation distance $T_\text{dist}$. 
We use MASt3R-SfM~\cite{DBLP:journals/corr/abs-2409-19152} to estimate the pose of the generated results.
The generated pose trajectories are aligned with the first frame of the groundtruth, then normalized to match the scale of the groundtruth trajectory.

\subsection{Sparse View Interpolation}
\label{sec_svi}
\noindent\textbf{Large Overlap.} We conduct quantitative comparisons on sparse input views. 
The competitors include NeRF-based methods~\cite{DBLP:conf/cvpr/YuYTK21,DBLP:conf/cvpr/XuCCSZP0024}, light field-based methods~\cite{suhail2022generalizable, DBLP:conf/cvpr/DuSTS23}, 3D-GS based methods~\cite{DBLP:conf/cvpr/CharatanLTS24,DBLP:journals/corr/abs-2403-14627,DBLP:conf/eccv/ZhangBTXZSX24,xu2024depthsplat}, and two concurrent works ReconX~\cite{DBLP:journals/corr/abs-2408-16767} and ViewCrafter~\cite{DBLP:journals/corr/abs-2409-02048} which are closely related to our approach as they both combine a large reconstruction model with a video diffusion model.
Note that except for ViewCrafter and ours, all other methods explicitly reconstruct the 3D representation and then render it to the novel views, while ViewCrafter and our approach directly generate images between the input images with pose control.
As listed in~\cref{tab:nvs}, both ReconX and ours are among the best.

\noindent\textbf{Small Overlap}. We further conduct qualitative comparisons on more challenging scenarios where the two input images have minimal or no overlap.
The competitors include the pure reconstruction-based MVSplat~\cite{DBLP:journals/corr/abs-2403-14627} and the combined reconstruction and generation-based ViewCrafter.
As shown in~\cref{fig:large_angle}, MVSplat struggles to inpaint invisible regions, resulting in holes or distortions.
ViewCrafter relies on the reconstructed point clouds from input images. When the overlap is too small, it generates many unreasonable contents or ghosting artifacts.
In contrast, our method can interpolate reasonable content between input images with precise pose control.

\subsection{Perpetual View Generation}
\label{sec_pvg}


\noindent\textbf{Short-Range Video}. 
We conduct quantitative comparisons on short-range video generation.
\rev{To ensure a fair comparison with ViewCrafter, which supports a maximum of 25 frames per clip, we uniformly evaluate the first generated 25 frames for all methods.}
As listed in~\cref{tab:Zero_Shot}, our method consistently achieves the best results in all metrics.
\rev{In short-range video generation, most of the content is derived from the first frame, with only a small portion being generated. Therefore, the high similarity scores primarily reflect the superior fidelity and pose accuracy of our method.}
Additionally, compared to RealEstate-10K, the videos in Tanks-and-Temples exhibit faster motion speeds, leading to varying degrees of performance decline across all methods.
Despite not being trained on this dataset, our model still achieves commendable metrics, demonstrating its strong generalization capability.



\noindent\textbf{Long-Range Video}. 
We compare the scalability on long-range videos.
Typically, as the generated video becomes longer, the fidelity tends to degrade, and the content gradually deviates from the input trajectory.
As shown in~\cref{fig:fid_long}, the scalability of our method significantly outperforms existing methods in terms of both fidelity (measured by FID) and pose accuracy (measured by $R_\text{dist}$ and $T_\text{dist}$).
The increase in these metrics by our method is significantly slower compared to other methods.
\rev{As illustrated in~\cref{fig:ar}, this trend is further confirmed.}
While ViewCrafter exhibits significant degradation as the generated video becomes longer, our method is able to generate reasonable content throughout the entire sequence.

\subsection{Layout-Conditioned City Generation}
\label{sec_lccg}


 We conduct qualitative comparison with CityDreamer~\cite{DBLP:conf/cvpr/Xie0H024}, a state-of-the-art method for layout-conditioned city generation. City maps are randomly sampled from OpenStreetMap~\cite{osmdataset}, and camera trajectories are randomly generated to render depth and semantic maps, as shown in the ``Layout'' column of~\cref{fig:layout}. Both methods can effectively follow the desired layout, but our method achieves significantly better fidelity.
 

\subsection{Ablation Study}
\label{sec_ablation}

\noindent\textbf{Spatiotemporal Autoregression}. We conduct ablation experiments on the task of perpetual view generation with 100-frame videos to evaluate the impact of our proposed spatial and temporal conditions.
As shown in~\cref{tab:Ablation_Study_AR}, the temporal conditions are crucial for maintaining fidelity, while the spatial conditions enable accurate pose control. When both components are combined in our method, there is a significant improvement in both visual fidelity and pose accuracy, underscoring the benefits of leveraging both spatial and temporal information together.

\noindent\textbf{Spatiotemporal-Conditioned Video Generation}. We conduct ablation experiments on the task sparse view interpolation with single-clip videos to evaluate the impact of each design choice in the proposed spatiotemporal-conditioned video generation, as shown in~\cref{tab:Ablation_Study_LRM}. ``w/ DUSt3R'' replaces the proposed LRM with DUSt3R, as done in ViewCrafter. ``w/o spatial cond.'' refers to a pure video interpolation method where only the input images are fixed, without the inclusion of LRM.  ``w/o depth input'' excludes depth maps predicted by Depth Anything V2 as input, while ``w/o depth loss'' omits the novel view depth loss. Finally, ``fix LRM'' involves freezing the LRM parameters in the third training stage, training only the ControlNet. The results indicate that excluding any of these design choices leads to performance degradation.
\section{Conclusion}


In this work, we propose a novel autoregression framework that combines both spatial and temporal conditions to support long-range scene generation with precise pose control.
The framework is used to implement three downstream tasks, including sparse view interpolation, perpetual view generation, and layout-conditioned city generation.
The quantitative and qualitative evaluations demonstrate that the proposed method achieves superior scalability, fidelity, and pose accuracy compared to state-of-the-art methods.

One limitation of our method is handling large loops.
Without absolute constraints, the content generated in the last clip before closing a loop might significantly differ from the content at the other end of the loop.
The subsequent clip that closes the loop will attempt to interpolate between these mismatched sections, leading to unrealistic and unreasonable results.
Additionally, we have not yet reconstructed the generated video into a 3D representation.
To fully cover a large-scale scene, the planned generation trajectory might need to contain complex loops, making the spatial consistency among the generated video clips crucial for 3D reconstruction.
These are areas for our future research.

\section*{Acknowledgment}
The authors would like to thank Tong He and Wanli Ouyang for their constructive discussions and kind help in the preliminary exploration in this research.


{
    \small
    \bibliographystyle{ieeenat_fullname}
    \bibliography{main}
}

\end{document}